\documentclass{article}

\PassOptionsToPackage{square,numbers}{natbib}

     \usepackage[preprint]{neurips_2023}

\usepackage[utf8]{inputenc} 
\usepackage[T1]{fontenc}    
\usepackage{hyperref}       
\usepackage{url}            
\usepackage{graphicx}
\usepackage{amsmath}
\usepackage{subfigure}
\usepackage{booktabs}       
\usepackage{amsfonts}       
\usepackage{nicefrac}       
\usepackage{microtype}      
\usepackage{xcolor}         
\usepackage{amssymb}
\usepackage{amsthm}
\usepackage{multirow}
\newtheorem{theorem}{Theorem}
\newtheorem{definition}{Definition}
\newtheorem{lemma}{Lemma}
\title{Dendritic Integration Based Quadratic Neural Networks Outperform Traditional Aritificial Ones}

\usepackage{authblk}
\author[a,b]{Chongming Liu}
\author[a,b,c,1]{Songting Li }
\author[a,b,c,d,1]{Douglas Zhou}
\affil[a]{School of Mathematical Sciences, Shanghai Jiao Tong University, Shanghai 200240, China}
\affil[b]{Institute of Natural Sciences, Shanghai Jiao Tong University, Shanghai 200240, China}
\affil[c]{Ministry of Education Key Laboratory of Scientific and Engineering Computing, Shanghai Jiao Tong University, Shanghai 200240,
China}
\affil[d]{Shanghai Frontier Science Center of Modern Analysis, Shanghai Jiao Tong University, Shanghai 200240, China}

\begin{document}

\maketitle

\footnotetext[1]{Corresponding author: songting@stju.edu.cn, zdz@sjtu.edu.cn} 
\begin{abstract}

 Incorporating biological neuronal properties into Artificial Neural Networks (ANNs) to enhance computational capabilities poses a formidable challenge in the field of machine learning. Inspired by recent findings indicating that dendrites adhere to quadratic integration rules for synaptic inputs, we propose a novel ANN model, Dendritic Integration-Based Quadratic Neural Network (DIQNN). This model shows superior performance over traditional ANNs in a variety of classification tasks. To reduce the computational cost of DIQNN, we introduce the Low-Rank DIQNN, while we find it can retain the performance of the original DIQNN. We further propose a margin to characterize the generalization error and theoretically prove this margin will increase monotonically during training. And we show the consistency between generalization and our margin using numerical experiments. Finally, by integrating this margin into the loss function, the change of test accuracy is indeed accelerated. Our work contributes a novel, brain-inspired ANN model that surpasses traditional ANNs and provides a theoretical framework to analyze the generalization error in classification tasks.

\end{abstract}

\section{Introduction}

While the artificial neural network (ANN) framework has made significant advancements towards solving complex tasks, it still faces problems that are rudimentary to real brains \cite{Chavlis2021DrawingIF}. A notable distinction between the modern ANN framework and the human brain is that the former relies on a significant number of training samples, which consumes large amounts of energy, whereas the latter runs on extremely low power (<20 watts) and possesses a strong generalization capability based on few-shot learning. Studies have demonstrated that incorporating dendritic features in ANNs can alleviate these issues and enhance overall performance \cite{Wu2018ImprovedET,Payeur2020BurstdependentSP,Liu2020DendriteNA}. However, it is difficult to quantify the nonlinear integration of dendrite which is an essential property that allows individual neurons to perform complex computations \cite{Spruston2008PyramidalND, Ujfalussy2018GlobalAM}. Consequently, simple nonlinear functions like ReLU and Sigmoid are often used in dendritic-inspired models\cite{Jones2021MightAS,Ojha2022BackpropagationNT}. Moreover, theoretical analysis is crucial for the understanding of how dendritic-inspired properties can enhance the performance of ANNs, e.g., small generalization error in classification tasks.  

To address these issues, we propose a novel ANN model, i.e., Dendritic Integration-Based Quadratic Neural Network (DIQNN), based on the recent studies indicating that the somatic response of biological neurons obeys a quadratic integration rule when multiple synaptic inputs are received on the dendrites \cite{Hao2009AnAR,Li2014BilinearityIS}. Our model replaces the linear integration and nonlinear activation function with a simple quadratic integration, as shown below:
\begin{equation}\label{DIQNN}
    f(x)=\sigma(w\cdot x+b)\rightarrow f(x)=x^TAx.
\end{equation}
Using various classification datasets, we demonstrate that DIQNN significantly improve classification performance compare to the traditional ANN. To further reduce the computational cost of DIQNN, we introduce the Low-Rank DIQNN, while it can maintain the performance of the original DIQNN. To theoretically understand why Low-Rank DIQNN performs well, we present a normalized margin to characterize the generalization error and prove that the margin increases monotonically during training process. We show the consistency between generalization and our margin on datasets such as MNIST\cite{LeCun2005TheMD}, FMNIST\cite{Xiao2017FashionMNISTAN} and CIFAR-10. Finally, by explicitly adding our margin into the loss function, we indeed observe a noticeable acceleration in the change of test accuracy in Low-Rank DIQNN.

The remainder of this paper is organized as follows. Section 2 provides an overview of previous works related to dendritic-inspired models, quadratic neural networks, and ANN classification margin. In Section 3, we present DIQNN's performance and the low-rank property in classification tasks. This motivates us to propose Low-Rank DIQNN in Section 4. We further propose a new margin and analyze the classification performance of Low-Rank DIQNN. Section 5 contains conclusions.

\section{Related Work}

\vspace{0.1in}
\textbf{Dendritic-inspired computational model.} It is widely acknowledged that dendrites play a pivotal role in the nonlinear integration of neural systems, enabling individual neurons to execute intricate tasks\cite{Spruston2008PyramidalND,Ujfalussy2018GlobalAM}. Recently, numerous studies have explored the implementation of dendritic features in ANNs from different aspects, yielding encouraging results. In \cite{Wu2018ImprovedET,Jones2021MightAS,Ojha2022BackpropagationNT}, they examined the incorporation of dendrite's morphology in ANNs and achieved higher test accuracy than traditional ANNs. Dendritic plasticity rule were used to design learning algorithms in \cite{Payeur2020BurstdependentSP,Moldwin2021TheGC}, to replace the non-biological backpropagation algorithm, and obtained enhanced performance on classification tasks. In \cite{Liu2020DendriteNA}, they represented dendritic nonlinearity by replacing the standard linear summation with polynomial summation and reported an improvement in approximation and classification experiments.  

\vspace{0.1in}
\textbf{Quadratic neural networks (QNNs).} Previous works have examined quadratic integration as an alternative to the traditional linear summation.  Table~\ref{quadratic-like-review} presents several commonly studied neuron models in quadratic formats and their corresponding references. One-rank format of quadratic neurons is provided in \cite{Goyal2020ImprovedPN,DeClaris1991ANC,Fan2017ANT,Bu2021QuadraticRN,Xu2022QuadraLibAP}, and enhanced performance on classification tasks is reported. In \cite{Jiang2019NonlinearCI,Mantini2021CQNNCQ,Zoumpourlis2017NonlinearCF}, quadratic neuron is only used in the convolution layer, leading to test accuracy improvements.  In \cite{Redlapalli2003DevelopmentOQ}, a single-layer network with quadratic format is applied to solve the XOR problem. In contrast, Table~\ref{quadratic-like-review} shows that our Low-Rank DIQNN approach totally differs from previous models, and we provide a theoretical framework for analyzing the generalization error of our model. 

\begin{table}[htbp!]
    \caption{The Overview of current QNN  works.}
    \footnotesize
    \label{quadratic-like-review}
    \centering
    \begin{tabular}{ccccc}
    \toprule
    Work reference &  Quadratic neuron format & \begin{tabular}{c}
          Layers for using\\
        quadratic neuron
    \end{tabular} & \begin{tabular}{c}
          Model\\
        structure \end{tabular} & \begin{tabular}{c}
             Theory for  \\
             generalization
        \end{tabular}\\
     \midrule
       \begin{tabular}{c}
           \cite{Goyal2020ImprovedPN},\cite{DeClaris1991ANC}, \cite{Fan2017ANT},   \\
             \cite{Bu2021QuadraticRN},\cite{Xu2022QuadraLibAP}
       \end{tabular} &  $f(x) = (w_a x)\cdot(w_b x)$ & \begin{tabular}{c}
            Convolution  \\
            $\&$ Classifier 
       \end{tabular}  & Various & N \\
      \midrule
      \cite{Redlapalli2003DevelopmentOQ} & $f(x)=x^TWx$ & Classifier & 1-layer &  N\\
      \midrule
      \cite{Jiang2019NonlinearCI},\cite{Mantini2021CQNNCQ},\cite{Zoumpourlis2017NonlinearCF} & $f(x)=x^TWx$ & Convolution & Various &  N\\
        \midrule 
        Our DIQNN & $f(x)=x^TWx$ & Classifier & Various & Y \\ 
        \midrule
         \begin{tabular}{c}
             Our Low-   \\
               Rank DIQNN
         \end{tabular} & $f(x)=\sum_i (w_ix)\cdot (w_ix)$ &  Classifier & Various & Y \\
        \bottomrule
    \end{tabular}
\end{table}

\vspace{0.1in}
\textbf{ANN classification margin.} Margin is a useful metric for describing both the generalization error and the robustness of models\cite{Sokoli2017RobustLM,MoosaviDezfooli2015DeepFoolAS} However, how to accurately calculate the margin of an ANN model remains an open question \cite{Guo2021RecentAI}. In \cite{Kobayashi2019LargeMI,Wu2019UnderstandingAR}, they define a margin and explicitly add it to the loss function, and achieve enhanced model performance (generalization error, robustness, etc.). Meanwhile, local linearization is used in \cite{Yan2018DeepDT,Sokoli2017RobustLM} to effectively calculate the classification margin. However, the relationship between margin and generalization is still lacking. In \cite{Lyu2019GradientDM}, they define a margin for homogeneous neural networks and provide theoretical proof for its monotonicity and optimality when the loss is sufficiently small. However, the results can not describe the relation between margin and generalization error during the early stages of training process.  To contribute to this area, we propose our margin and theoretically prove this margin increases throughout the training process. And numerical experiments show the consistency of dynamics between test accuracy and our margin.

\section{Our dendritic integration based quadratic neural network (DIQNN)}

All of the numerical experiments in this paper are conducted using Python and executed on a Tesla A100 computing card with a 7nm GA100 GPU, featuring 6,912 CUDA cores and 432 tensor cores.

\subsection{Performance of DIQNN on classification tasks}

Here, we primarily evaluate DIQNN's performance on two datasets: MNIST and FMNIST. (Supplementary materials include information on experimental setup and results for additional datasets such as Iris.)

\begin{table}[htbp!]
    \caption{Performance of different model on MNIST.}
    \label{qnn_compare}
    \centering
    \begin{tabular}{cccc}
    \toprule
       Network structure & \begin{tabular}{c}
       Linear net \\(784-10)\end{tabular} & \begin{tabular}{c}
     Linear net\\ (784-ReLU(8000)-10)  \end{tabular} & 
     Single layer DIQNN\\
     \midrule
       \begin{tabular}{c}
        Number of \\ trainable parameters \end{tabular} & 7.84*1e3 & 6.35*1e6 & 6.15*1e6 \\
        \midrule
        Test accuracy  &  89.8$\pm$0.5$\%$   & 93.1$\pm$0.4$\%$  &   \textbf{98.0$\pm$0.1$\%$}      \\
        \bottomrule
    \end{tabular}
\end{table}

\begin{figure}[htbp!]
\centering
    \subfigure{\includegraphics[width=0.4\textwidth]{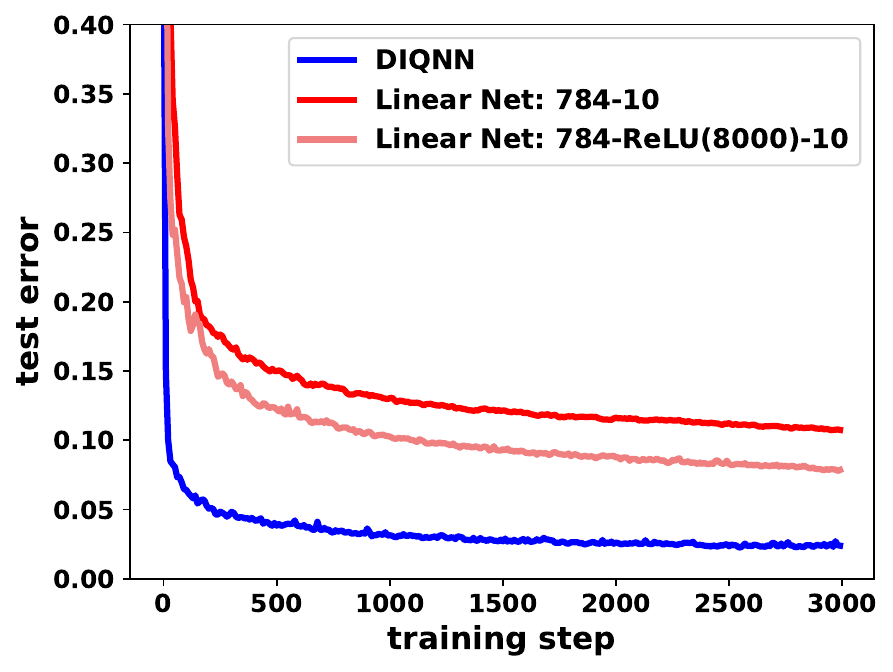}} 
    \subfigure{\raisebox{0.1\height}{\includegraphics[width=0.3\textwidth]{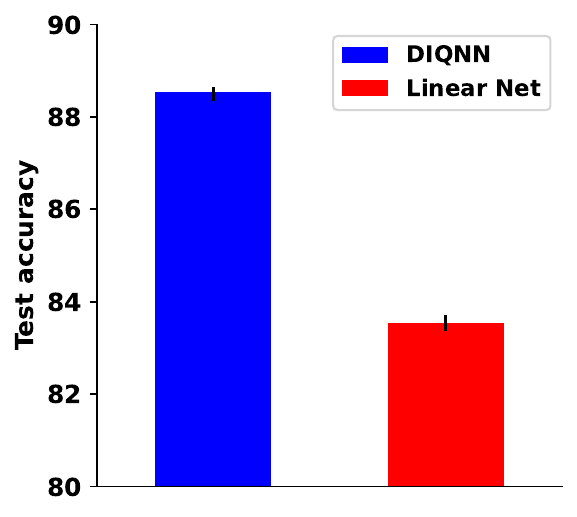}}}
    \caption{Performance of DIQNN. Left: Test error curves on MNIST. Right: Test accuracy (with error bar) on FMNIST.}
    \label{qnn_performance}
\end{figure}

\vspace{0.1in}
\textbf{MNIST results.} Our study evaluates single layer DIQNN, single layer linear net, and two-layer linear net with a nonlinear activation function ReLU (equation\ref{DIQNN}, left side). The test error curves are presented in the left figure of Figure\ref{qnn_performance}. Additionally, Table~\ref{qnn_compare} shows the number of trainable parameters and final test accuracy for each model. Our results indicate that DIQNN performs significantly better than linear net, even with almost identical number of trainable parameters. These findings underscore the advantages of DIQNN.

\vspace{0.1in}
 \textbf{FMNIST results.} We employ a two-layer convolutional neural network (CNN), comprising of one convolution layer with ReLU activation function and one fully connected layer. We examine whether the classifier ( the fully connected layer) is linear or quadratic. The test accuracy of these two models is presented in the right figure of Figure~\ref{qnn_performance}, indicating that DIQNN significantly outperforms the linear net.

\subsection{Low rank properties of DIQNN}
Although quadratic classifiers offer advantages, their computational cost is large. Therefore, it is imperative to reduce the number of trainable parameters while preserving the quadratic form in DIQNN. To this end, we want to find out whether the trained network can possess some good properties. Firstly, the weight matrix $A$ undergoes spectrum decomposition. The decomposition can be represented as follows\footnote{As $A$ is initialized by a symmetric matrix, it is always symmetric with gradient descent algorithm.}:
\begin{equation}\label{eigen_decomposition}
    A=\sum_{i=1}^n\lambda_i\mathbf{a}_i\mathbf{a}_i^T,\ |\lambda_1|\geq|\lambda_2|\geq|\lambda_3|\geq\cdots\geq|\lambda_n| 
\end{equation}

\begin{figure}[htbp!]
    \centering
    \subfigure{\includegraphics[width=0.19\textwidth]{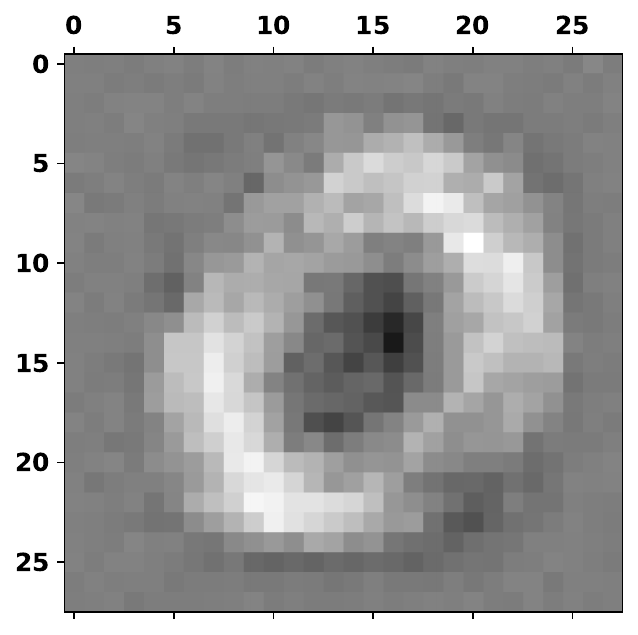}} 
    \subfigure{\includegraphics[width=0.19\textwidth]{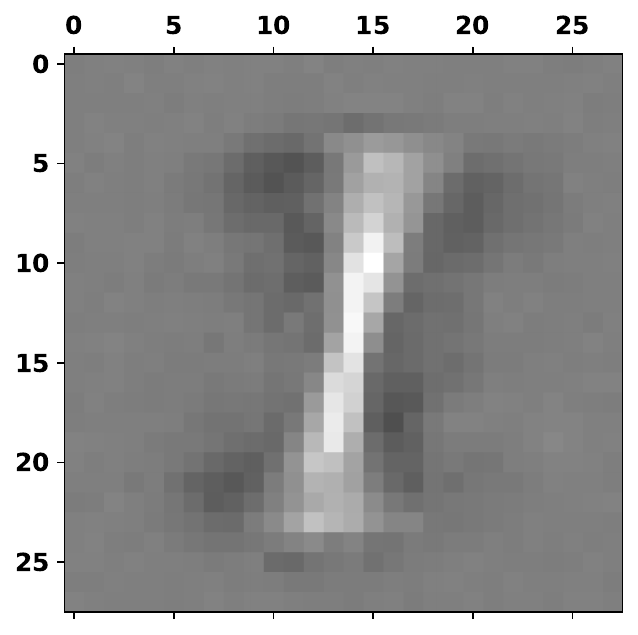}}
    \subfigure{\includegraphics[width=0.19\textwidth]{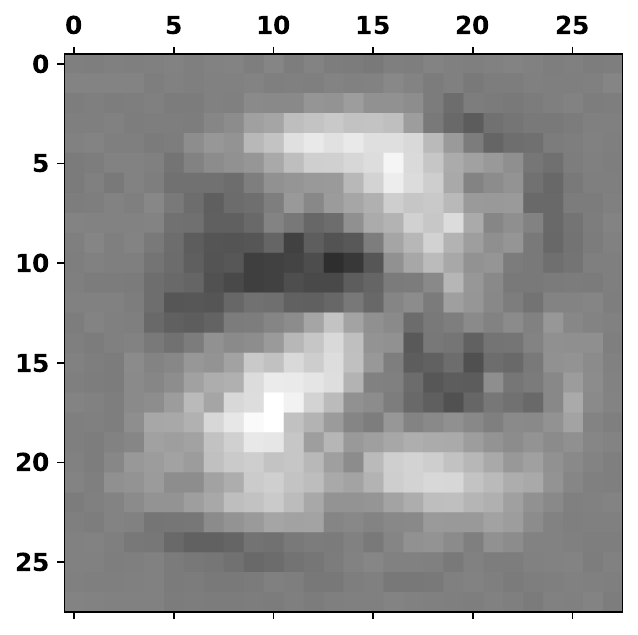}}
    \subfigure{\includegraphics[width=0.19\textwidth]{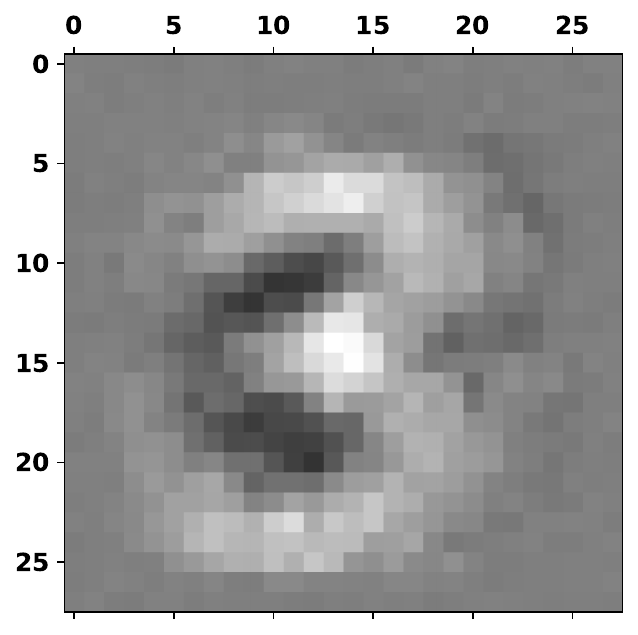}}
    \subfigure{\includegraphics[width=0.19\textwidth]{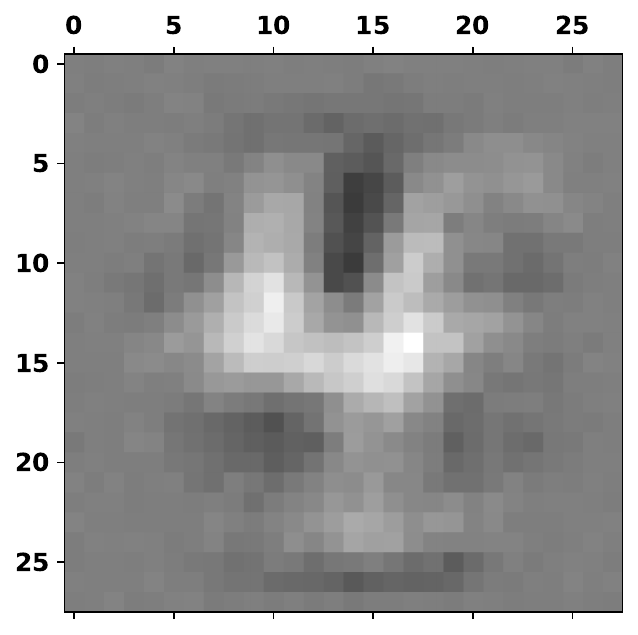}}
    \subfigure{\includegraphics[width=0.19\textwidth]{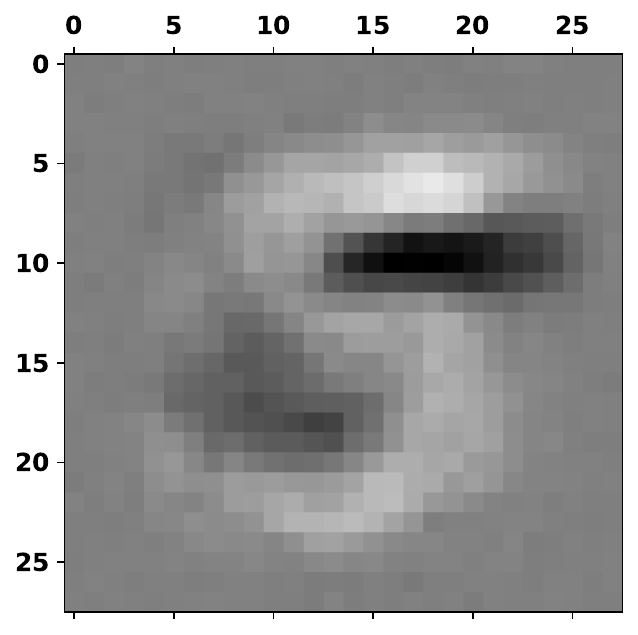}}
    \subfigure{\includegraphics[width=0.19\textwidth]{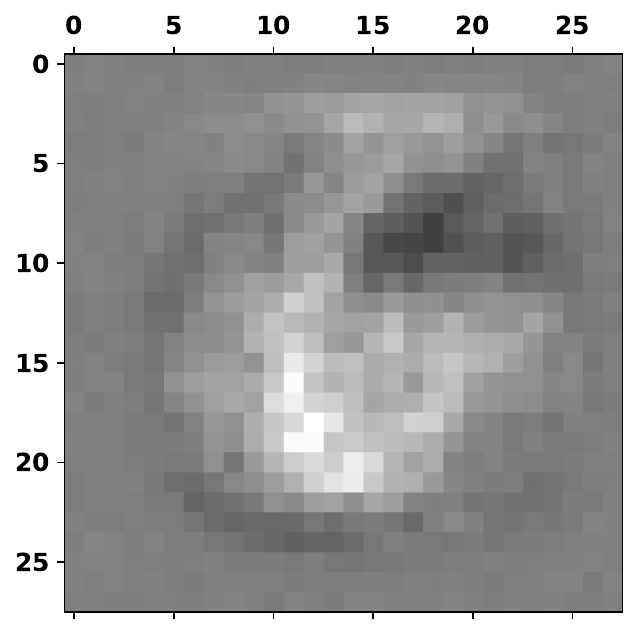}}
    \subfigure{\includegraphics[width=0.19\textwidth]{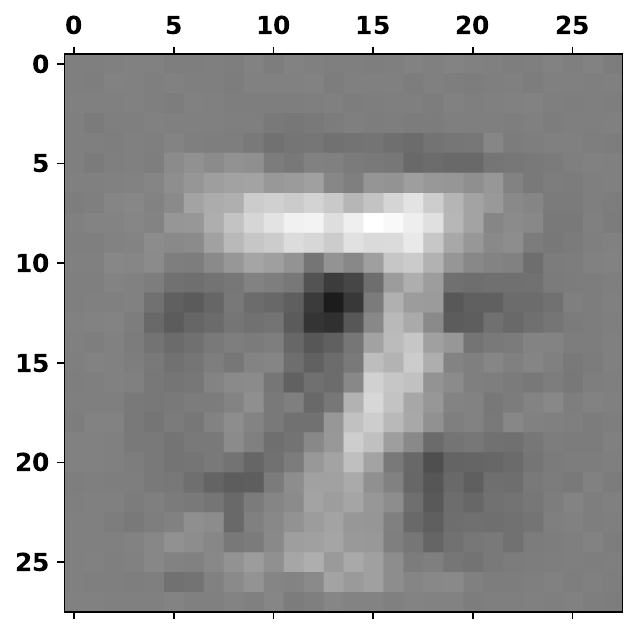}}
    \subfigure{\includegraphics[width=0.19\textwidth]{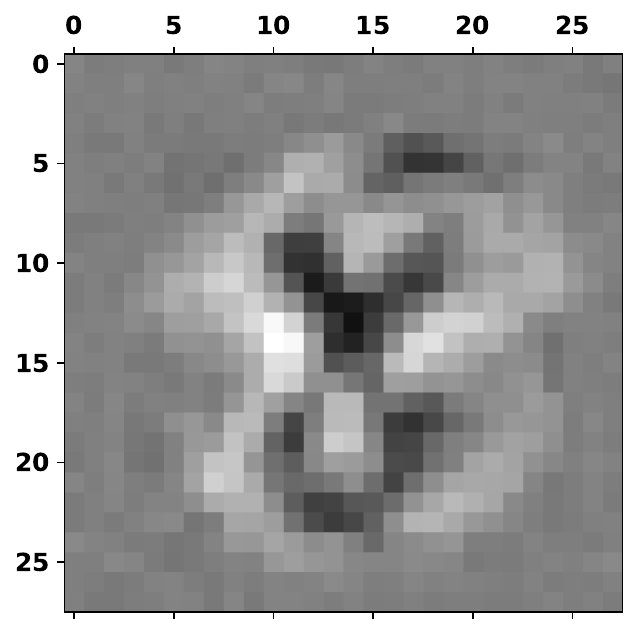}}
    \subfigure{\includegraphics[width=0.19\textwidth]{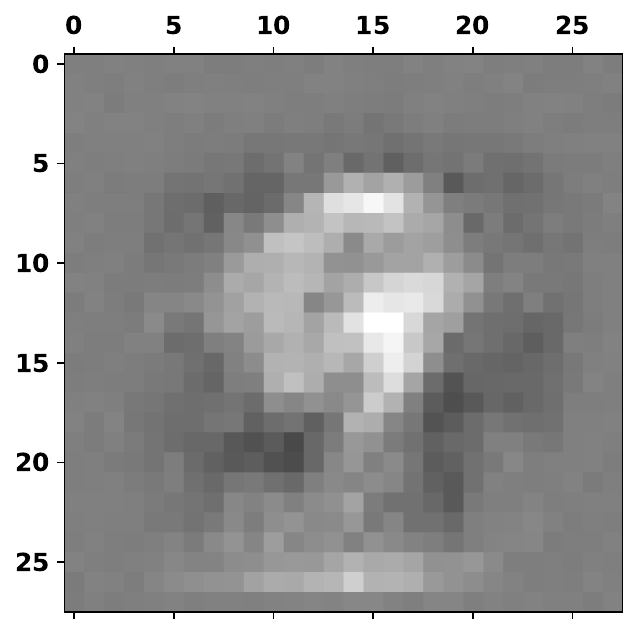}}
    \caption{The leading eigenvectors of different after-trained weight matrices.(Totally 10 weight matrices)}
    \label{low_rank_structure_in_qnn}
\end{figure}

Here, numerical experiment is conducted on MNIST. Figure~\ref{low_rank_structure_in_qnn} illustrates the leading eigenvectors of different after-train weight matrices. It can be observed that these vectors are similar to the spike-triggered average of output neurons. The term "spike-triggered average" refers to the average input that evokes the highest response from a neuron\cite{Schwartz2006SpiketriggeredNC}. This phenomenon implies the majority of information in spike-triggered average could be encoded by the leading eigenvector of the weight matrix $A$. In addition, we find when only the first few eigenvectors are kept in weight matrix $A$, DIQNN can achieve high test accuracy. Experiments on FMNIST exhibit similar results. Additional details are provided in supplementary materials. These observations suggest that DIQNN naturally display low-rank properties after training.

\section{Low-Rank DIQNN}
 The low-rank property of DIQNN allows us to explicitly set the weight matrix $A$ in the form of a sum of rank-one matrices given by the following equation: 
\begin{equation}\label{low rank DIQNN}
    A=\sum_{i=1}^{r}\mathbf{c}_i\mathbf{c}_i^T
\end{equation}
where $r$ is a hyperparameter and $\mathbf{c}_i$ is a column vector. We refer to this model as Low-Rank DIQNN owing to its capability of reducing computational cost, since $r$ is always small in comparison to the dimension of weight matrix $A$.

\subsection{Performance of Low-Rank DIQNN on classification tasks}

We next investigate the performance of Low-Rank DIQNN on MNIST, FMNIST, and CIFAR-10. The details of the numerical experiments and results on other datasets such as Iris are provided in supplementary materials.

\vspace{0.1in}
\textbf{MNIST and FMNIST results.} We utilize the same network structure as in the previous section for Low-Rank DIQNN. On MNIST, we employ single-layer Low-Rank DIQNNs with various ranks, while on FMNIST, we utilize a two-layer convolutional neural network that contains one low-rank classifier.

\begin{figure}[htbp!]
\centering
    \subfigure{\includegraphics[width=0.45\textwidth]{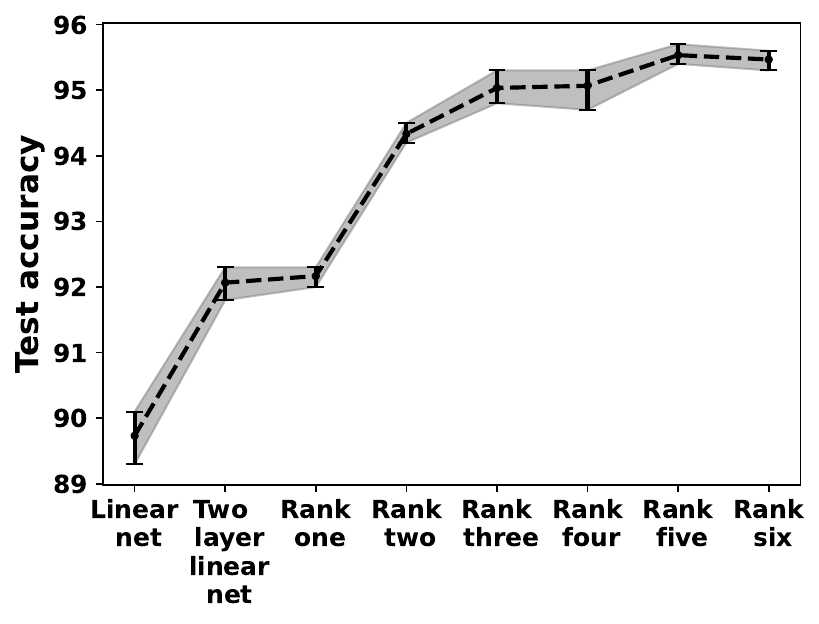}} 
    \subfigure{\raisebox{0.1\height}{\includegraphics[width=0.47\textwidth]{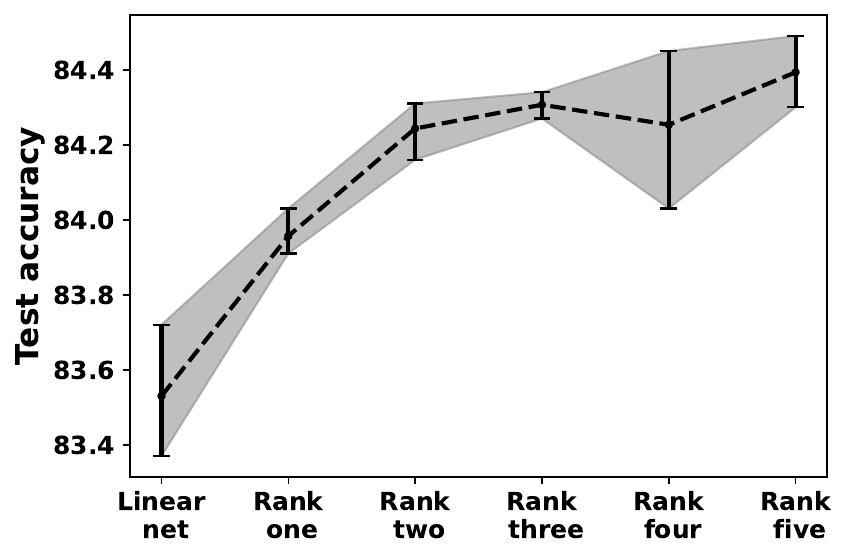}}}
    \caption{Performance of Low-Rank DIQNN. Left: On MNIST. Right: On FMNIST.}
    \label{lrnn_performance}
\end{figure}

Figure~\ref{lrnn_performance} mainly compares linear net with Low-Rank DIQNN having various ranks. The figure shows that when Low-Rank DIQNN with just one rank ($r=1$ in equation~\ref{low rank DIQNN}) performs better than the linear net, even though they have an equal number of trainable parameters (Low-Rank DIQNN with one rank even outperforms a two-layer linear net with ReLU activation function). Moreover, as the number of ranks in Low-Rank DIQNN increases, its performance continues to improve until it almost reaches saturation.

\begin{table}[htbp!]
    \caption{Performance of Low-Rank DIQNN on CIFAR-10.}
    \scriptsize
    \label{lrnn_compare_cifar10}
    \centering
    \begin{tabular}{cccc}
    \toprule
       Network structure &  Residual blocks & Last layer method (Classifier)  & 
     Test accuracy\\
     \midrule
       ResNet 10 & [1,1,1,1] & \begin{tabular}{c}
            Linear  \\
             Rank one \\
             Rank two \\
             Rank three \\
             Rank four \\
       \end{tabular} & \begin{tabular}{c}
            92.20$\pm$0.11$\%$  \\
             92.32$\pm$0.18$\%$ \\
             92.35$\pm$0.09$\%$ \\
             92.48$\pm$0.07$\%$ \\
             \textbf{92.50$\pm$0.11$\%$} \\
       \end{tabular} \\
        \midrule
        ResNet 12 & [2,1,1,1] & \begin{tabular}{c}
            Linear  \\
             Rank one \\
             Rank two \\
             Rank three \\
             Rank four \\
       \end{tabular} & \begin{tabular}{c}
            92.59$\pm$0.14$\%$  \\
             92.63$\pm$0.15$\%$ \\
             92.76$\pm$0.08$\%$ \\
             92.72$\pm$0.15$\%$ \\
             \textbf{92.89$\pm$0.14$\%$} \\
       \end{tabular} \\
        \midrule
        ResNet 14 & [2,2,1,1] & \begin{tabular}{c}
            Linear  \\
             Rank one \\
             Rank two \\
             Rank three \\
             Rank four \\
       \end{tabular} & \begin{tabular}{c}
            92.82$\pm$0.20$\%$  \\
             93.00$\pm$0.07$\%$ \\
             92.89$\pm$0.17$\%$ \\
             93.30$\pm$0.12$\%$ \\
             \textbf{93.31$\pm$0.19$\%$} \\
       \end{tabular} \\
        \midrule
        ResNet 16 & [2,2,2,1] & \begin{tabular}{c}
            Linear  \\
             Rank one \\
             Rank two \\
             Rank three \\
             Rank four \\
       \end{tabular} & \begin{tabular}{c}
            93.27$\pm$0.08$\%$  \\
             93.26$\pm$0.04$\%$ \\
             93.29$\pm$0.05$\%$ \\
             93.33$\pm$0.17$\%$ \\
             \textbf{93.43$\pm$0.05$\%$} \\
       \end{tabular} \\
        \midrule
        ResNet 18 & [2,2,2,2] & \begin{tabular}{c}
            Linear  \\
             Rank one \\
             Rank two \\
             Rank three \\
             Rank four \\
       \end{tabular} & \begin{tabular}{c}
            93.51$\pm$0.04$\%$  \\
             93.46$\pm$0.13$\%$ \\
             93.57$\pm$0.03$\%$ \\
             \textbf{93.67$\pm$0.13$\%$}\\
             93.66$\pm$0.06$\%$ \\
       \end{tabular} \\
        \bottomrule
    \end{tabular}
\end{table}

\vspace{0.1in}
\textbf{CIFAR-10 results.} We train ResNets\cite{He2015DeepRL} with a low-rank classifier on CIFAR-10. Similar to the CNN used in FMNIST, we keep the computation of convolution layers in ResNet intact, and only modify the last fully connected layer to low-rank form. Test accuracy results on different ResNets are presented in Table~\ref{lrnn_compare_cifar10}, where the "Residual blocks" column means how these ResNets are constructed in terms of their convolution layers \cite{He2015DeepRL}. Similar to the results on MNIST and FMNIST, it is evident that Low-Rank DIQNN performs better than traditional ANN, and its performance increases with higher ranks.

\subsection{Analyzing generalization of Low-Rank DIQNN on classification tasks}

This section begins by defining the margin on Low-Rank DIQNN. We then utilize it to analyze Low-Rank DIQNN and show the consistency between generalization and our margin. Finally, we show this margin can be used to accelerate the change of test accuracy in Low-Rank DIQNN. The details of the numerical experiments and the proof of our theorem are provided in supplementary materials.

\subsubsection{Normalized margin}
Given training dataset $\{(x_n,y_n)\}_{n=1}^N$, each data point $x_n\in \mathbb{R}^d$ has a true label $y_n\in [k]$, where $k$ ($k\geq 2$) denotes the total number of classes. Given a certain input $x_n$, the Low-Rank DIQNN produces an output vector $\Phi(x_n,\theta)\in \mathbb{R}^{k}$, where $\theta$ represents the trainable network parameters. Naturally, the margin for data point $x_n$ can be defined as $s_{n}=(\Phi(x_n,\theta))_{y_n}-(\Phi(x_n,\theta))_{j}, \ j=\arg\max_{i\neq y_n} {(\Phi(x_n,\theta))_i}$. However, this definition does not directly reflect the distance between data point $x_n$ and the classification boundary of the model in input space since it does not obey scale invariance. To address this issue, we propose a normalized margin for single data point $x_n$: $\frac{s_n}{\|\Phi(x_n,\theta)\|_2}$, and the average of this margin across all training data points are taken into consideration:
\begin{equation}\label{our margin}
    \frac{1}{N}\sum^{N}_{n=1}\frac{s_{n}}{\|\Phi(x_n,\theta)\|_2} = \frac{1}{N}\sum^{N}_{n=1}\frac{(\Phi(x_n,\theta))_{y_n}}{\|\Phi(x_n,\theta)\|_2}-\frac{1}{N}\sum^{N}_{n=1}\ \frac{(\Phi(x_n,\theta))_{j}}{\|\Phi(x_n,\theta)\|_2}\triangleq\mu_1-\mu_2\triangleq\Delta\mu
\end{equation}

\subsubsection{XOR problem}
We employ a single-layer Low-Rank DIQNN with one rank to solve the XOR problem and prove the following theorem (see details in supplementary materials):
\begin{theorem}
Training single layer Low-Rank DIQNN with one rank to solve XOR problem under cross-entropy loss $\mathcal{L}(\theta_t)$ and gradient flow algorithm as $\frac{d\theta_t}{dt}=-\nabla_\theta \mathcal{L}(\theta_t)$, we have:
\begin{equation*}
    \frac{d\Delta\mu}{dt}>0\ \  \text{and}\ \lim_{t\rightarrow\infty}\Delta\mu=1.
\end{equation*}
\end{theorem}
 With our defined margin, the above theorem demonstrates that our margin will monotonically increase and finally reach its maximum value. Notably, in Low-Rank DIQNN, one can prove $-1\leq\Delta\mu\leq 1$.

\begin{figure}[htbp!]
\centering
    \subfigure{\includegraphics[width=0.19\textwidth]{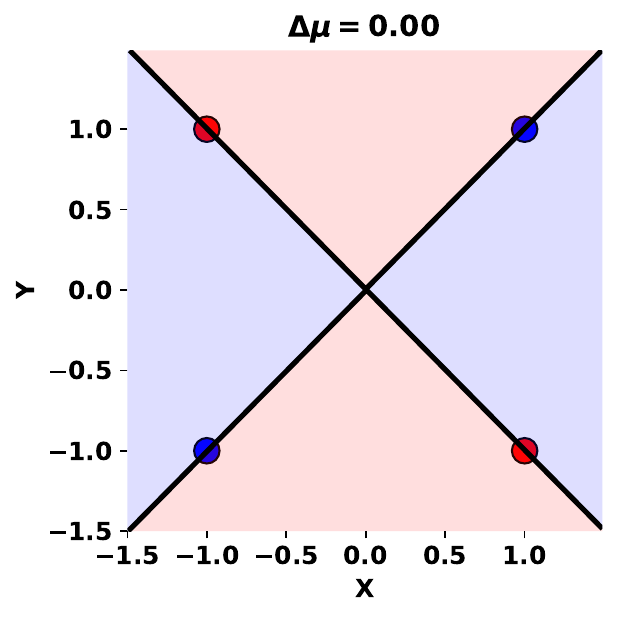}} 
    \subfigure{\includegraphics[width=0.19\textwidth]{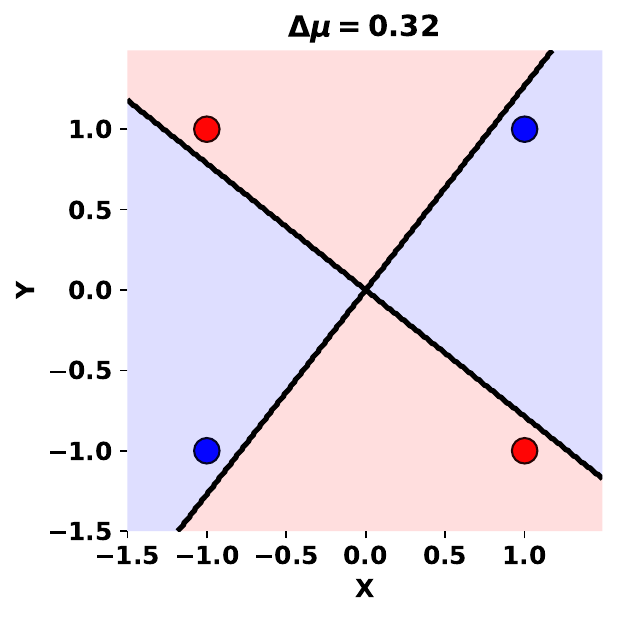}} 
    \subfigure{\includegraphics[width=0.19\textwidth]{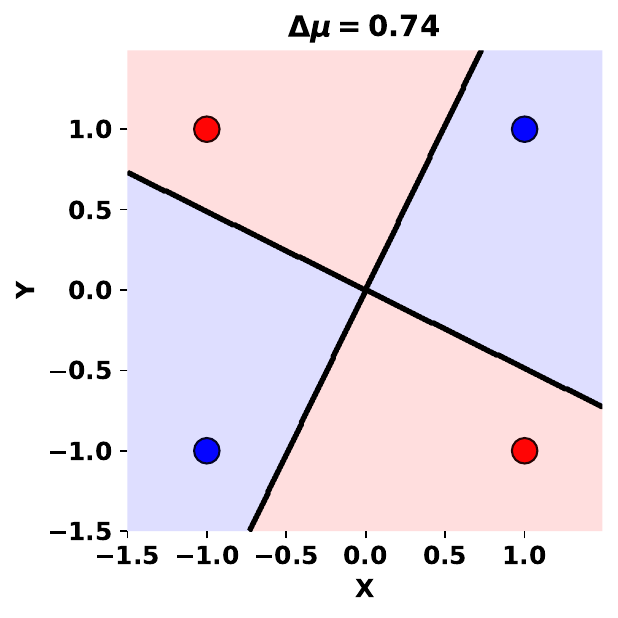}} 
    \subfigure{\includegraphics[width=0.19\textwidth]{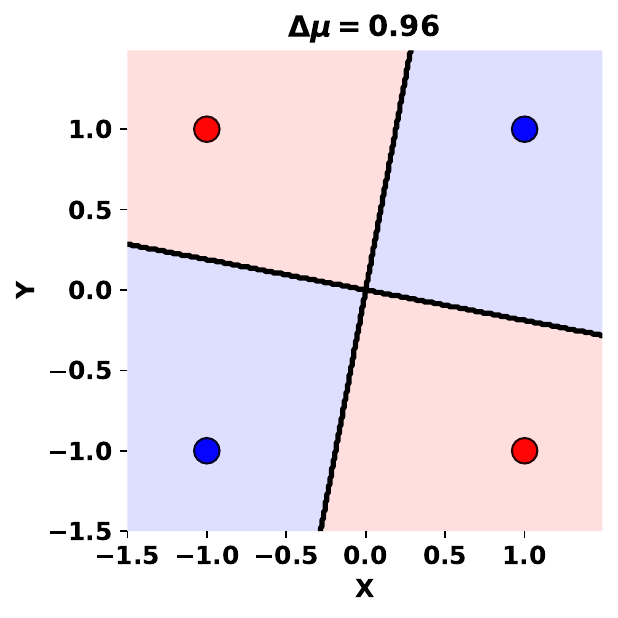}} 
    \subfigure{\includegraphics[width=0.19\textwidth]{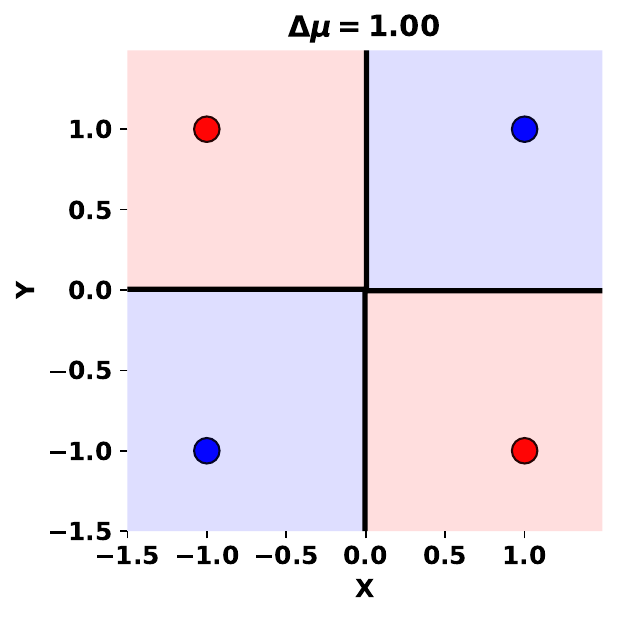}} 
    \caption{Training process (gradient descent) of single layer Low-Rank DIQNN with one rank on XOR problem. From left to right: Initialization, after 12, 36, 84, 396 training steps, respectively. The blue region indicates the model's classification for the blue class, while the red region indicates classification for the red class. And the title of each picture indicates the value of our margin $\Delta\mu$ at different training steps.}
    \label{xor}
\end{figure}

We perform numerical experiments, as shown in Figure~\ref{xor}. These figures display the training dynamics with gradient descent (GD). It shows that our margin reflects the relative distance between data points and the classification boundary. And we observe that Low-Rank DIQNN can achieve a smaller generalization error by pushing the classification boundary further away from data points until it converges to an optimal result, which aligns with our theorem. Therefore, our theorem implies that Low-Rank DIQNN reduces generalization error monotonically until converging to the optimal solution.

\subsubsection{General set up}

We next extend our investigation to general datasets and training algorithms (e.g., stochastic gradient descent (SGD)).

\begin{definition}(Homogeneous property)
     We say a network satisfy the homogeneous property for some integer $L$ if:
     \begin{equation*}
         \Phi(x_n,\theta_t)=\|\theta_t\|_2^L\Phi(x_n,\frac{\theta_t}{\|\theta_t\|_2})
     \end{equation*}
\end{definition}

\begin{lemma}(Homogeneous property)
 Low-Rank DIQNN satisfy the homogeneous property: $\Phi(x_n,\theta_t)=\|\theta_t\|_2^L\Phi(x_n,\frac{\theta_t}{\|\theta_t\|_2})$ for $L=2^{l+1}-2$, where $l$ is the total number of layers of Low-Rank DIQNN. Moreover, we can prove that $\langle\partial_\theta(\Phi(x_n,\theta_t))_i,\theta_t\rangle=L(\Phi(x_n,\theta_t))_i$  $\forall i\in [k]$
\end{lemma}

\begin{definition} 
     A set that consist of real numbers $\{y_1,y_2,\cdots,y_m\}$ satisfy the $\varepsilon-$separated condition if $y_j-\min_i (y_i)\geq 1/\varepsilon, \forall j
     \in [m], j\neq \arg\min_i (y_i)$
\end{definition}

\begin{definition}
      $s_{nj}=(\Phi(x_n,\theta_t))_{y_n}-(\Phi(x_n,\theta_t))_{j}$; $j_n = \arg\max_{i\neq y_n} {(\Phi(x_n,\theta))_i}$; $s_{n}=s_{nj_{n}}$; $c$ is the condition number of matrix $A=\partial_\theta S\partial_\theta S^T-\frac{L^2}{\|\theta_t\|_2^2}SS^T$, where $S=(s_1,\dots,s_n)^T$, $\partial_\theta S=(\partial_\theta s_1, \dots, \partial_\theta s_n)^T$; $\mathbf{v}=(v_1,\dots,v_N)^T$, where $v_n=\frac{e^{(\Phi(x_n,\theta_t))_{j_n}}}{\sum_{i=1}^k e^{(\Phi(x_n,\theta_t))_i}}$, it should be noted that all of these values or vectors are time dependent.
\end{definition}

\begin{theorem}
   Training Low-Rank DIQNN under cross-entropy loss and gradient flow algorithm, if the following three assumptions are satisfied in a small time interval $[t-\Delta t,t+\Delta t]$ for some $\Delta t>0$:
   \begin{itemize}
       \item $\|\Phi(x_n,\theta_t)\|_2=a_n\|\theta_t\|_2^L$
       \item The set $\{s_{nj}\}_{j=1,j\neq y_n}^{k}$ satisfy the $\varepsilon-$separated condition for some $\varepsilon>0$ ($\forall n\in [N]$), and $\{\|\partial_\theta s_{nj}\|_2\}_{n\in [N], j\in [k]}$ is uniformly bounded with some constant $M$.
       \item $c-1 \leq \frac{2m}{\sqrt{1-m^2}}$, where $m=\cos(\mathbf{a},\mathbf{v})>0$, $\mathbf{a}=(1/a_1,\dots,1/a_N)^T$
   \end{itemize}
   Then at time $t$ we have:
   \begin{equation*}
    \frac{d\Delta\mu}{dt}\geq -\frac{M^2(k-2)\|\mathbf{a}\|_1}{N\|\theta_t\|_2^L}e^{-\frac{1}{\varepsilon}}.
\end{equation*}
\end{theorem}

\begin{figure}[htbp!]
\centering
    \subfigure{\includegraphics[width=0.32\textwidth]{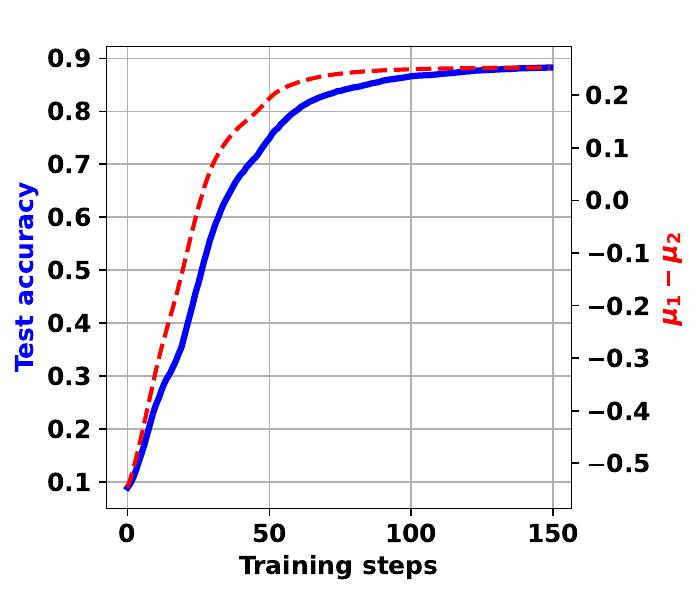}} 
    \subfigure{{\includegraphics[width=0.32\textwidth]{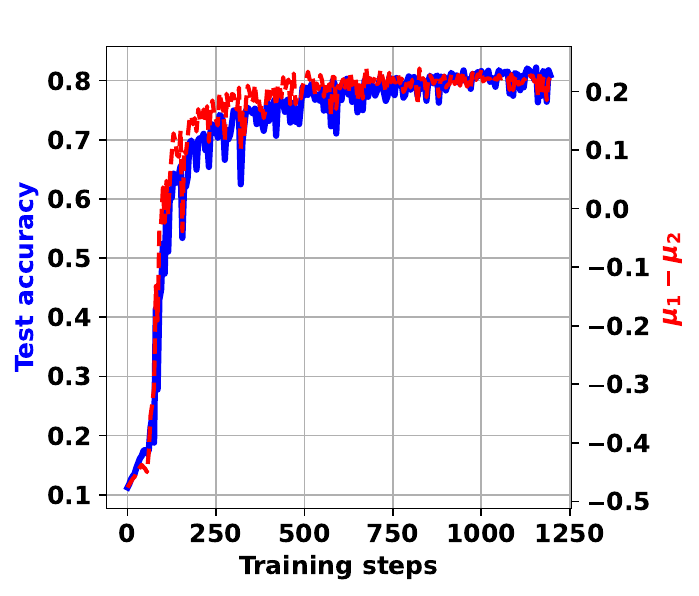}}}
    \subfigure{{\includegraphics[width=0.32\textwidth]{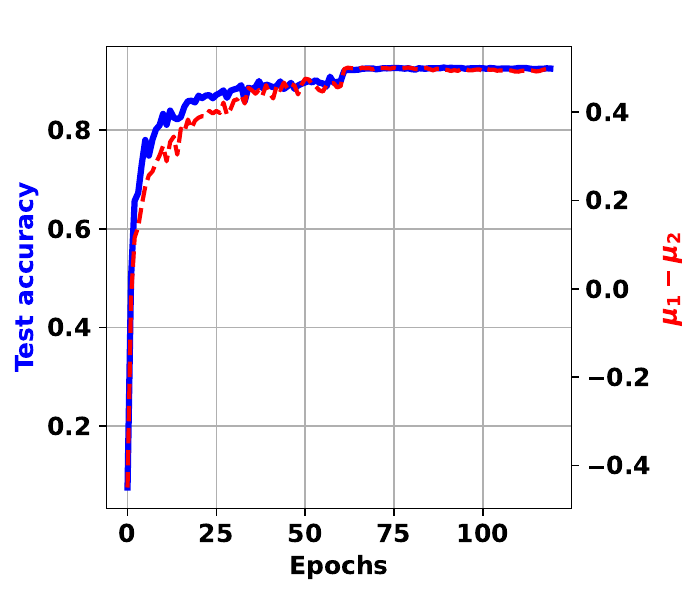}}}
    \caption{Numerical experiments on MNIST(Left), FMNIST(Middle) and CIFAR-10(Right). There are two vertical axes: the left axis represents the test accuracy, and the right axis represents our margin $\Delta\mu$. The blue solid line indicates test accuracy while the red dash line indicates our margin.}
    \label{margin_validity}
\end{figure}

This theorem indicates that through the training process, our defined margin increases in an almost monotonic manner\footnote{In fact, this theorem is also valid for Low-Rank DIQNN with SGD, the original DIQNN with both GD and SGD. See supplementary materials for details.}

We next perform numerical experiments on MNIST, FMNIST, and CIFAR-10 with SGD training algorithm, as shown in Figure~\ref{margin_validity}. These figures demonstrate that the margin increases almost monotonically on these datasets, which is consistent with our theorem. Additionally, it is noteworthy that the dynamics of the margin is closely related to that of test accuracy, indicating that the margin well reflects the generalization error of Low-Rank DIQNN. Our theorem implies that the generalization error decreases monotonically, indicating the Low-Rank DIQNN is approaching the solution with good generalization.

\begin{figure}[htbp!]
\centering
    \subfigure{\includegraphics[width=0.32\textwidth]{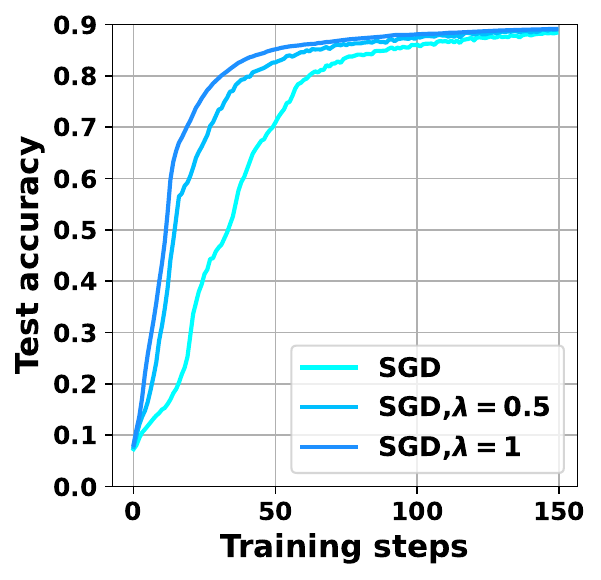}} 
    \subfigure{{\includegraphics[width=0.32\textwidth]{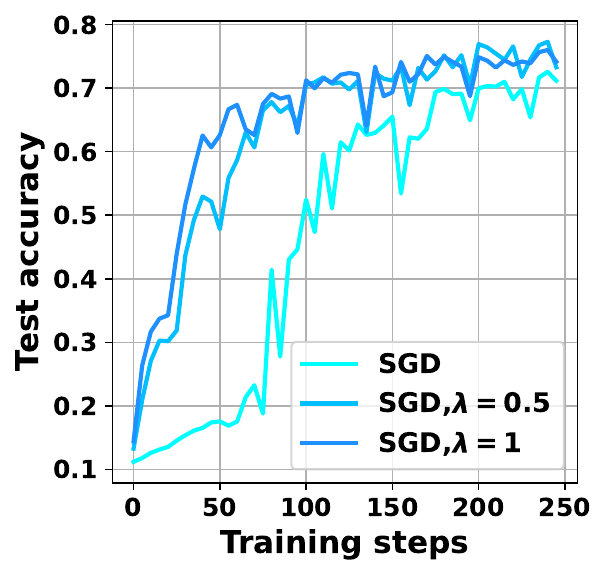}}}
    \subfigure{{\includegraphics[width=0.32\textwidth]{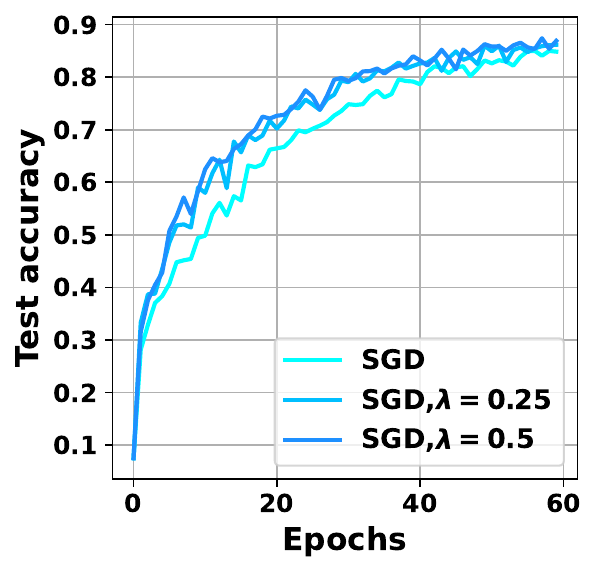}}}
    \caption{Test accuracy curves on datasets: MNIST(Left), FMNIST(Middle) and CIFAR-10(Right), respectively, with loss function given in equation~\ref{loss_regularization}.}
    \label{regularization}
\end{figure}

The monotonical increase of our margin during training process may imply the existence of implicit regularity. Therefore, we next explore the impact of explicitly adding margin regularization to the loss function. The modified loss function is given as follows:
\begin{equation}\label{loss_regularization}
    \bar{\mathcal{L}}(\theta)=\mathcal{L}(\theta)-\lambda*\Delta\mu,
\end{equation}
where $\mathcal{L}(\theta)$ is the original cross-entropy loss, and $\lambda$ is a hyperparameter that controls the magnitude of regularization. Numerical results on MNIST, FMNIST, and CIFAR-10 are provided in Figure~\ref{regularization}. It is found that the convergence speed is indeed enhanced after the margin regularization is added, here the convergence speed is characterized by the test accuracy, since one cares more about the generalization capability of solutions. This indicates one can obtain a solution with small generalization error in a short time by explicitly adding margin regularization.

\section{Conclusion}

In this paper, we present a novel ANN model inspired by the quadratic integration rule of dendrite, namely DIQNN, which significantly outperforms traditional ANNs. Based on observations of the low-rank structure of DIQNN, we propose Low-Rank DIQNN that retains the high performance while having low computational cost. Moreover, we provide a theoretical framework to analyze the generalization error in classification tasks, and experimental results on MNIST, FMNIST, and CIFAR-10 exhibit the effectiveness of our framework. Considering quadratic integration rule of neurons could be confined to a few brain areas\cite{Hao2009AnAR}, thus other integration rules of neurons may be discovered in future electrophysiological experiments. Our framework could be extended to design brain-inspired ANN models that can incorporate these new integration rules. Moreover, in the design of brain-inspired deep neural networks, it might be possible to refer to our framework to incorporate different integration rules in different layers, which represent neurons in various brain areas. Additionally, how to theoretically analyze the above brain-inspired models developed from new integration rules will also be an important issue. It is expected to extend our margin theory to investigate the generalization error of these new brain-inspired models in classification tasks.

\medskip

\bibliography{ref}


\end{document}


\maketitle

\section{Dendritic integration based quadratic neural network (DIQNN)}

The form of DIQNN discussed in the main text are given as follows:
\begin{equation}\label{DIQNN}
    f(x)=x^TAx
\end{equation}

\subsection{Performance of DIQNN on other datasets}

\begin{table}[htbp!]
    \caption{Datasets' information.}
    \label{toy_dataset}
    \centering
    \begin{tabular}{cccc}
    \toprule
       \textbf{Name} & \textbf{Feature dimension} & \textbf{Number of classes}  & 
     \textbf{Dataset size}\\
     \midrule
       \textbf{Iris} & 4 & 3 & 150 \\
        \midrule
        \textbf{Wine}  &  13   & 3  &   178     \\
                \midrule
        \textbf{Wiscosin}  &  30   & 2  &  569     \\
                \midrule
       \textbf{Handwritten digit}  &  64   & 10 &   1794     \\
        \bottomrule
    \end{tabular}
\end{table}

\begin{figure}[htbp!]
    \centering
    \includegraphics[width=0.95\textwidth]{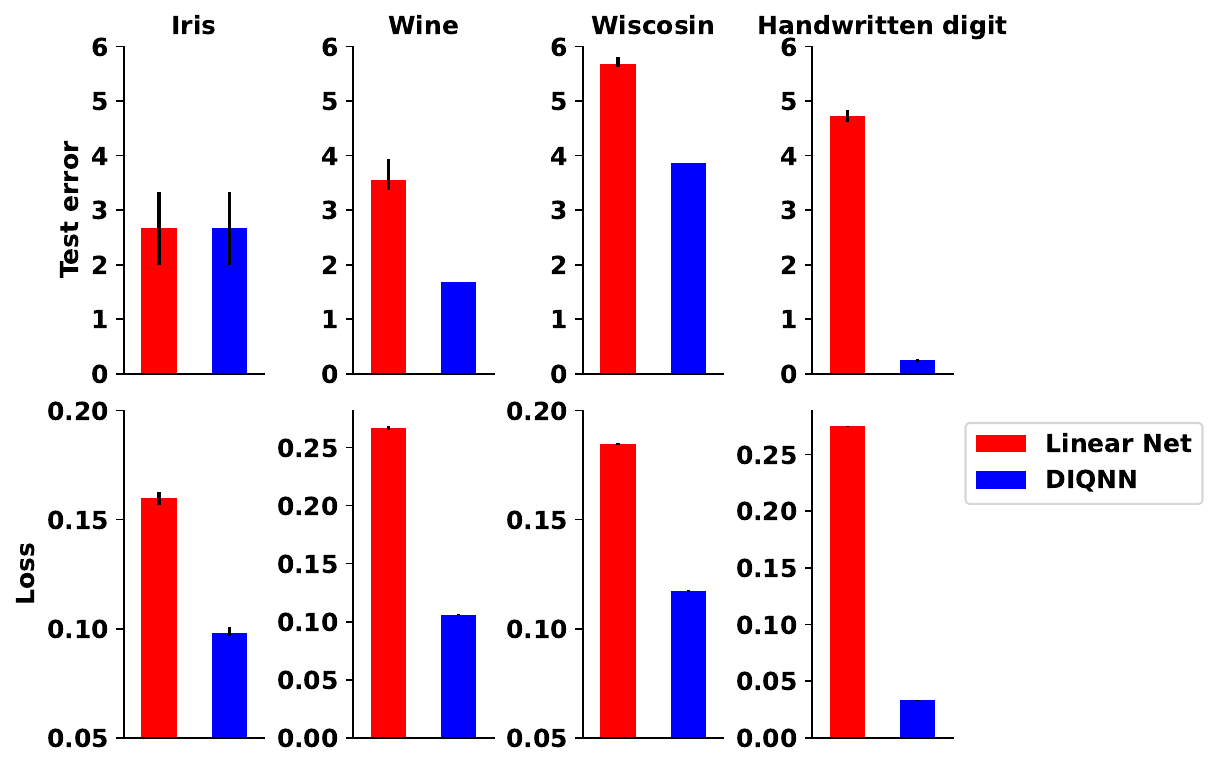}
    \caption{Performance of DIQNN on Iris, Wine, Wisconsin and Handwritten digit datasets.}
    \label{toy_dataset_qnn}
\end{figure}

Here we conduct numerical experiments on four classification datasets: Iris, Wine, Wisconsin and Handwritten digit, all these datasets are available at \url{http://archive.ics.uci.edu/ml}. Information about these four datasets are given in Table~\ref{toy_dataset}. And here we randomly choose some of the data points to form test set, and the remaining data points are treated as train set. Then we train a single layer linear net and a single layer DIQNN under cross-entropy loss with gradient descent algorithm , and compare the final results under the same choice of hyperparameters as shown in Table~\ref{experiment_set_up}.

Performance is given in Figure~\ref{toy_dataset_qnn}. These results indicate that DIQNN outperforms the linear net, in terms of both generalization (test error) and training loss.

\subsection{Low rank properties of DIQNN}

\begin{figure}[htbp!]
    \centering
    \includegraphics[width=\textwidth]{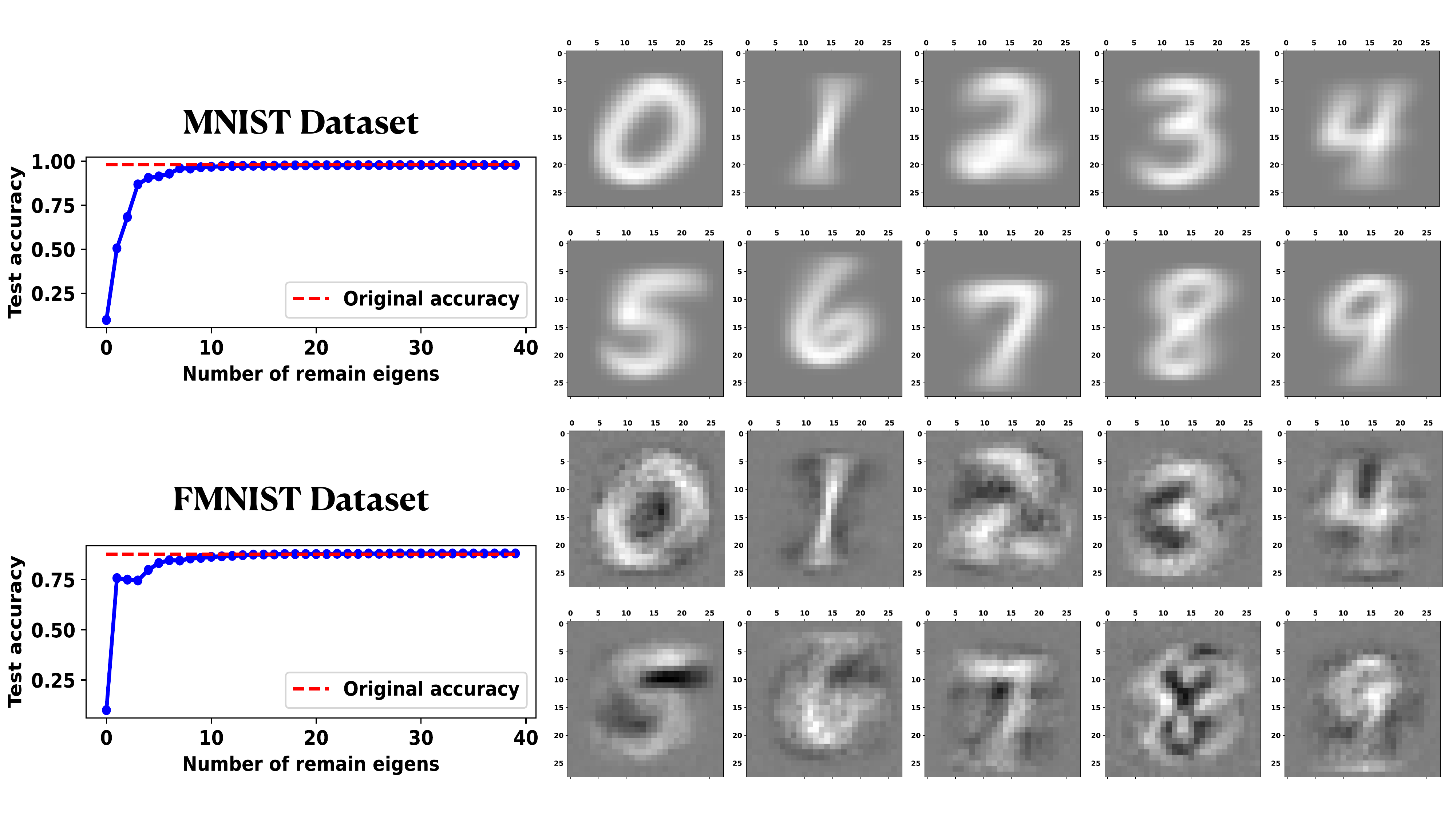}
    \caption{ Low rank properties in DIQNN. Upper left: The relationship between test accuracy and the number of retained eigenvectors sorted in descending order on MNIST. Lower left: The same as upper left, but on FMNIST dataset. Upper right (first two rows of 10 images): Spike-triggered average of the output neurons (totally 10 output neurons). Lower right (last two rows of 10 images, the figure in the main text): The leading eigenvectors of different weight matrices (totally 10 weight matrices).}
    \label{low_rank_property}
\end{figure}

In the main text, we have discussed the low rank properties of DIQNN from two aspects: (a) the relationship between test accuracy and the number of retained eigenvectors sorted in descending order demonstrates that the original test accuracy can be achieved when the first few eigenvectors are kept in weight matrix $A$; (b) the similarity between the leading eigenvectors of weight matrices and the spike-triggered average of output neurons indicates that the majority of information in spike-triggered average could be encoded by the leading eigenvector of the weight matrix $A$. Here, the spike-triggered average is computed by averaging the inputs that are classified into the same category. These results are shown in Figure~\ref{low_rank_property}.

\section{Experimental set up}

The experimental details, i.e. the choice of hyperparameters, are given in Table~\ref{experiment_set_up}, and the comparison between linear net and our models (DIQNN and Low-Rank DIQNN) is under the same choice of hyperparameters. It should be noted that on CIFAR-10 dataset, we use a weight decay of 0.001 and momentum of 0.9, and start with a learning
rate of 0.002, divide it by 10 at 60 epochs and terminate training at 250 epochs.

\begin{table}[htbp!]
    \caption{Experiment details.}
    \small
    \label{experiment_set_up}
    \centering
    \begin{tabular}{cccc}
    \toprule
       \textbf{Dataset} & \textbf{Learning rate} & \textbf{Batch size}  & 
     \textbf{Number of epochs}\\
     \midrule
       \begin{tabular}{c}
           \textbf{Iris, Wine,}   \\
             \textbf{Wiscosin, Handwritten digit}
       \end{tabular} & 0.01 & - (Full batch) & 1000 \\
        \midrule
        \textbf{MNIST}  &  0.01   & 100  &   5     \\
                \midrule
        \textbf{FMNIST}  &  0.01   & 100  &  20     \\
                \midrule
       \textbf{CIFAR-10}  &  0.002   & 128 &   250     \\
        \bottomrule
    \end{tabular}
\end{table}

\section{Low-Rank DIQNN}

The form of weight matrix in Low-Rank DIQNN discussed in the main text is given as follows:
\begin{equation}\label{low rank DIQNN}
    A=\sum_{i=1}^{r}\mathbf{c}_i\mathbf{c}_i^T
\end{equation}

\subsection{Performance of Low-Rank DIQNN on other datasets}

\begin{figure}[htbp!]
    \centering
    \subfigure{\includegraphics[width=0.4\textwidth]{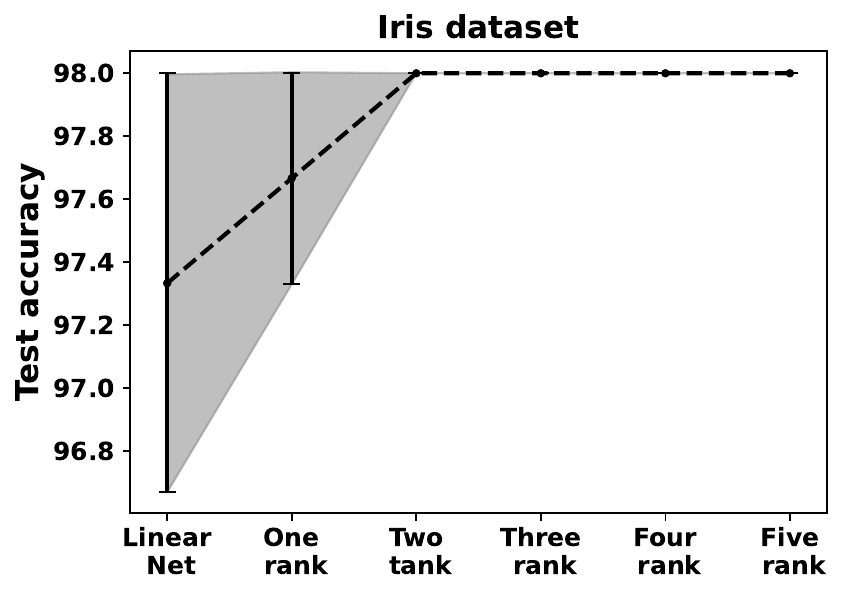}} 
    \subfigure{\includegraphics[width=0.4\textwidth]{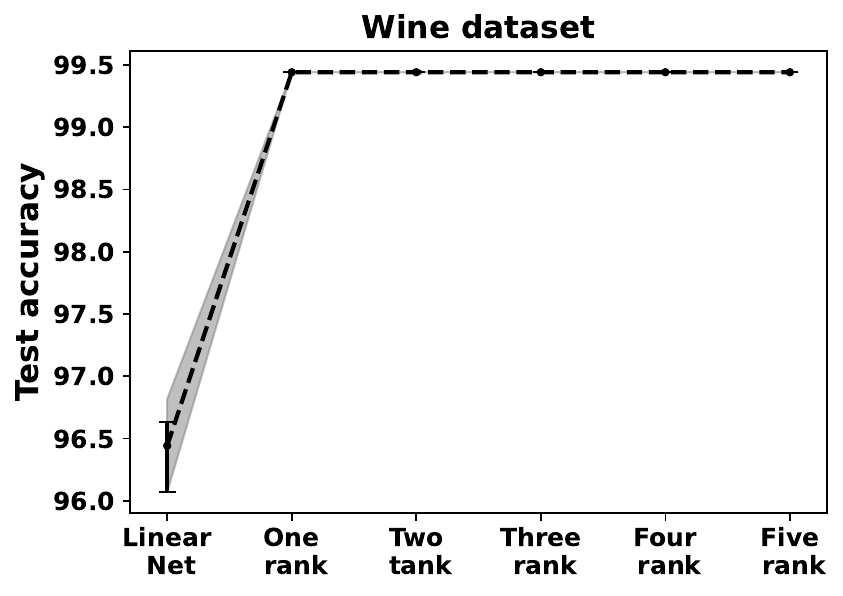}}
    \subfigure{\includegraphics[width=0.4\textwidth]{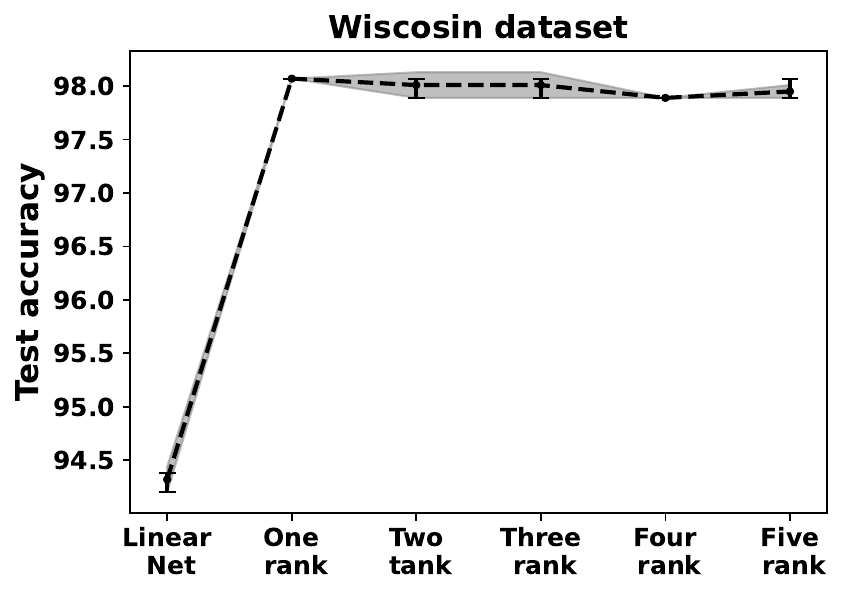}}
    \subfigure{\includegraphics[width=0.4\textwidth]{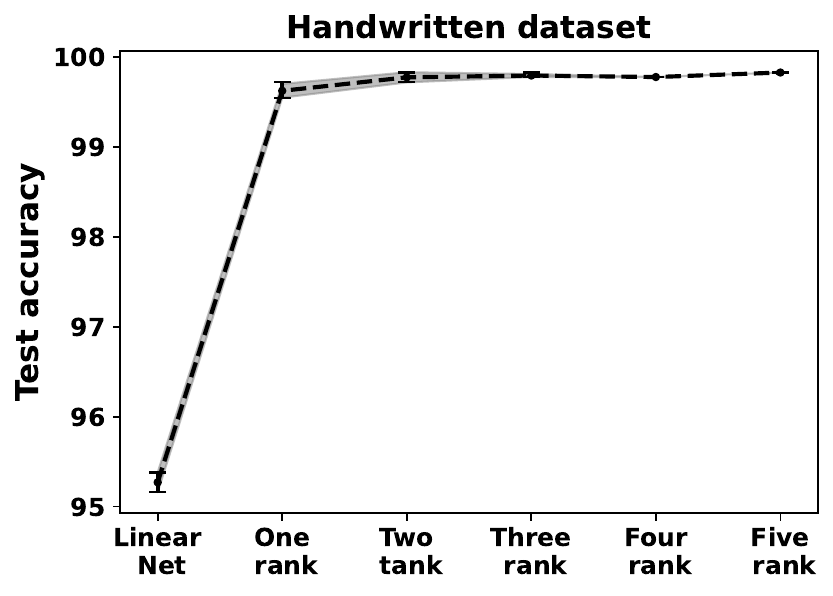}}
    \caption{Performance of Low-Rank DIQNN on Iris, Wine, Wisconsin and Handwritten digit datasets.}
    \label{toy_dataset_lrnn}
\end{figure}

Here, we conduct numerical experiments on Iris, Wine, Wiscosin and Handwritten digit. And the training details are given in Table~\ref{experiment_set_up}. Figure~\ref{toy_dataset_lrnn} shows the performance of Low-Rank DIQNN. Low-Rank DIQNN with one rank outperforms linear net on these datasets and the test accuracy is high and almost saturated when the weight matrix is rank one.

\subsection{Proof of the theorem on XOR problem}

\begin{theorem}
Using single layer Low-Rank DIQNN with one rank under cross-entropy loss $\mathcal{L}(\theta)$ and gradient flow algorithm on XOR problem\footnote{Here we only consider the intialization: $c_1=[1,0]^T, c_2=[0,1]^T$. }, we have:
\begin{equation*}
    \frac{d\Delta\mu}{dt}> 0\ \  \text{and}\ \lim_{t\rightarrow\infty}\Delta\mu=1.
\end{equation*}
\end{theorem}

\begin{proof}
\textbf{Idea:} Here we first try to reduce these four data points to one. After we have proved that we only need to consider a single data point, we can prove the result by explicitly writing down $\frac{d\Delta\mu}{dt}$.

\begin{figure}[htbp!]
    \centering
    \includegraphics[width=0.25\textwidth]{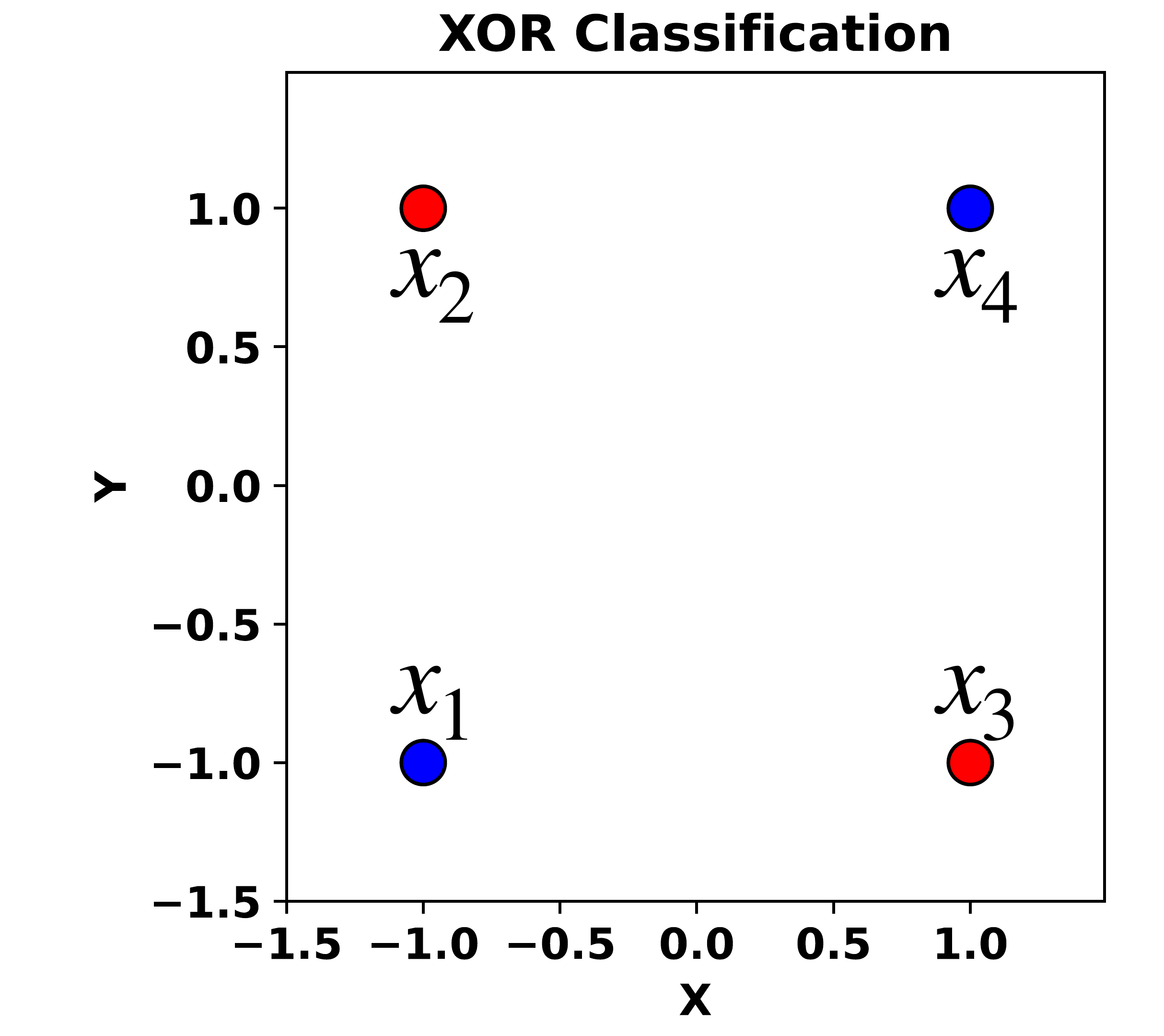}
\end{figure}

Here the network mapping is: $\Phi(\cdot,\theta): x\in\mathbb{R}^2\rightarrow (\langle c_1,x\rangle^2,\langle c_2,x\rangle^2)^T\in\mathbb{R}^2$, where $\theta = (c_1,c_2)$ is the trainable parameters of Low-Rank DIQNN. Due to the fact that $x_1=-x_4, x_2=-x_3$, The cross-entropy loss can be written as:
\begin{equation*}
    L(c_1,c_2)=(\log(1+e^{-s_{1}})+\log(1+e^{-s_{2}}))/2,
\end{equation*}
where $s_1 = \langle c_1,x_1 \rangle^2-\langle c_2,x_1\rangle^2, s_2=\langle c_2,x_2\rangle^2-\langle c_1,x_2\rangle^2$. Then by gradient flow algorithm, we can explicitly write down the dynamics of $\theta$:
\begin{equation*}
    \frac{dc_1}{dt}=-\frac{\partial L(c_1,c_2)}{\partial s_{12}}\langle c_1,x_1\rangle x_1+\frac{\partial L(c_1,c_2)}{\partial s_{21}}\langle c_1,x_2\rangle x_2,
\end{equation*}
\begin{equation*}
    \frac{dc_2}{dt}=-\frac{\partial L(c_1,c_2)}{\partial s_{21}}\langle c_2,x_2\rangle x_2+\frac{\partial L(c_1,c_2)}{\partial s_{12}}\langle c_2,x_1\rangle x_1.
\end{equation*}
Then we have: 
\begin{equation*}
    \frac{d\langle c_1,x_1\rangle ^2}{dt}=2v_{12}\langle c_1,x_1\rangle ^2, \ 
    \frac{d\langle c_2,x_1\rangle ^2}{dt}=-2v_{12}\langle c_2,x_1\rangle ^2,
\end{equation*}
where $v_{12}=\frac{1}{1+e^{-s_{1}}}, v_{21}=\frac{1}{1+e^{-s_{2}}}$. Then we have:
\begin{equation*}
    \frac{d(\langle c_1,x_1\rangle ^2*\langle c_2,x_1\rangle ^2)}{dt}=0\Rightarrow \langle c_1,x_1\rangle ^2*\langle c_2,x_1\rangle ^2=constant.
\end{equation*}
So $\langle c_1,x_1\rangle ^2, \langle c_2,x_2\rangle ^2$ satisfy the same dynamic system: 
\begin{equation*}
    \frac{df(t)}{dt}=2f(t)*\frac{1}{1+e^{f(t)-\frac{1}{f(t)}}}.
\end{equation*}
Under our set up, $\langle c_1,x_1\rangle ^2, \langle c_2,x_2\rangle ^2$ also share the same initial condition, then by the uniqueness and existence of ODE with a Lipshitz continuous condition, we have: $\langle c_1,x_1\rangle ^2 = \langle c_2,x_2\rangle ^2$. Similarily, we can prove that $\langle c_2,x_1\rangle^2=\langle c_1,x_2\rangle^2$. Then the explicit form of our margin can be written as:
\begin{equation*}
    \Delta\mu = \frac{a-b}{\sqrt{a^2+b^2}},
\end{equation*}
where $a=\langle c_1,x_1\rangle^2, b=\langle c_1,x_2\rangle^2$. Then computing $\frac{d\Delta\mu}{dt}$ explicitly:
\begin{equation*}
    \frac{d\Delta\mu}{dt}=\frac{4v_{12}(a+b)ab}{(a^2+b^2)\sqrt{a^2+b^2}}> 0.
\end{equation*}
Since at initialization $a=b=1\Rightarrow \lim\limits_{t\rightarrow\infty}a=+\infty, \lim\limits_{t\rightarrow\infty}b=0\Rightarrow \lim\limits_{t\rightarrow\infty}\Delta\mu=1$.

\end{proof}

\subsection{Proof of the theorem under general set up}

Here we try to understand our margin in a more general set up, which includes multi-layer Low-Rank DIQNN, general datasets and algorithms such as stochastic gradient descent. (SGD) 

Given training dataset $\{(x_n,y_n)\}_{n=1}^N$, which consists of $N$ data points. Each data point $x_n\in \mathbb{R}^d$ has a true label $y_n\in [k]$, where $k$ ($k\geq 2$) denotes the total number of classes. Given a certain input $x_n$, the network produces an output vector $\Phi(x_n,\theta)\in \mathbb{R}^{k}$, where $\theta$ represents the trainable network parameters.

\begin{definition}(Homogeneous property)
     We say a network satisfy the homogeneous property for some integer $L$ if:
     \begin{equation*}
         \Phi(x_n,\theta_t)=\|\theta_t\|_2^L\Phi(x_n,\frac{\theta_t}{\|\theta_t\|_2}).
     \end{equation*}
\end{definition}

\begin{lemma}(Homogeneous property)
 Low-Rank DIQNN satisfy the homogeneous property: $\Phi(x_n,\theta_t)=\|\theta_t\|_2^L\Phi(x_n,\frac{\theta_t}{\|\theta_t\|_2})$ for $L=2^{l+1}-2$, where $l$ is the total number of layers of Low-Rank DIQNN. Moreover, we can prove that $\langle\partial_\theta(\Phi(x_n,\theta_t))_i,\theta_t\rangle=L(\Phi(x_n,\theta_t))_i$  $\forall i\in [k]$.
\end{lemma}
\begin{proof}
    Due to the fact that Low-Rank DIQNN does not have bias term, and the nonlinearity is square, it is obvious that $\exists L$ such that $\Phi(x_n,a\theta_t)=a^L\Phi(x_n,\theta_t)$. And if the total number of layers of Low-Rank DIQNN is $l$, we can write down the relation between $l$ and $L$:
    \begin{equation*}
        L=2^{l+1}-2.
    \end{equation*}
   If we take the derivative w.r.t $a$ on both sides in the equation: $(\Phi(x_n,a\theta_t))_i=a^L(\Phi(x_n,\theta_t))_i, \forall a\in\mathbb{R}$, we have:
    \begin{equation*}
    \langle\partial_\theta(\Phi(x_n,a\theta_t))_i,\theta_t\rangle=La^{L-1}(\Phi(x_n,a\theta_t))_i.
    \end{equation*}
    If we set $a=1$ on both side, then we have proved this lemma.
\end{proof}

\begin{definition}
      $s_{nj}=(\Phi(x_n,\theta_t))_{y_n}-(\Phi(x_n,\theta_t))_{j}$; $j_n = \arg\max_{i\neq y_n} {(\Phi(x_n,\theta))_i}$; $s_{n}=s_{nj_{n}}$; $c$ is the condition number of matrix $A=\partial_\theta S\partial_\theta S^T-\frac{L^2}{\|\theta_t\|_2^2}SS^T$, where $S=(s_1,\dots,s_n)^T$, $\partial_\theta S=(\partial_\theta s_1, \dots, \partial_\theta s_n)^T$; $\mathbf{v}=(v_1,\dots,v_N)^T$, where $v_n=\frac{e^{(\Phi(x_n,\theta_t))_{j_n}}}{\sum_{i=1}^k e^{(\Phi(x_n,\theta_t))_i}}$, it should be noted that all of these values or vectors are time dependent.
\end{definition}

\begin{definition} 
     A set that consist of real numbers $\{y_1,y_2,\cdots,y_m\}$ satisfy the $\varepsilon-$separated condition if $y_j-\min_i (y_i)\geq 1/\varepsilon, \forall j
     \in [m], j\neq \arg\min_i (y_i)$.
\end{definition}

Before proving the theorem, we firstly prove two lemmas which characterize the property of Low-Rank DIQNN and gradient flow algorithm with Crossentropy loss, respectively.

\begin{lemma}(Approximate $\frac{d\theta_t}{dt}$)
 If the set $\{s_{nj}\}_{j=1,j\neq y_n}^{k}$ satisfy the $\varepsilon-$separated condition for some $\varepsilon>0$ ($\forall n\in [N]$), and $\{\|\partial_\theta s_{nj}\|_2\}_{n\in [N], j\in [k]}$ is uniformly bounded with some constant $M$. $\forall n, \forall j\in [k]$  then under gradient flow dynamics with cross-entropy loss, we have: 
    \begin{equation*}
        \|\frac{d\theta_t}{dt}-\frac{1}{N}\sum_{n=1}^N v_n\partial_\theta s_n\|_2\leq M(k-2)e^{-\frac{1}{\varepsilon}}.
    \end{equation*}
\end{lemma}

\begin{proof}
    we can define $v_{nj}=\frac{e^{(\Phi(x_n,\theta_t))_j}}{\sum_{i=1}^k e^{(\Phi(x_n,\theta_t))_i}}$, which is the $j$-th softmax output when given input $x_n$ ($v_n=v_{nj_n}$). Then the crossentropy loss can be written as: $\mathcal{L}(\theta_t)=\frac{1}{N}\sum\limits_{n=1}^N\log(1+\sum\limits_{j\neq y_n}e^{-s_{nj}})$. Then by the gradient flow dynamics ($\frac{d\theta_t}{dt}=-\nabla_\theta\mathcal{L}(\theta_t)$), we have: $\frac{d\theta_t}{dt}=\frac{1}{N}\sum\limits_{n=1}^N\sum\limits_{j\neq y_n} v_{nj}\partial_\theta s_{nj}$. Then we can know:
    \begin{equation*}
    \|\frac{d\theta_t}{dt}-\frac{1}{N}\sum_{n} v_n\partial_\theta s_n\|_2=\frac{1}{N}\|\sum\limits_{n=1}^N\sum\limits_{j\neq y_n,j_n} v_{nj}\partial_\theta s_{nj}\|_2.
    \end{equation*}
    Due to the $\varepsilon-$separated condition, we have for $\forall n, \forall j\neq y_n,j_n$: 
    \begin{equation*}
        \frac{v_n}{v_{nj}} = e^{s_{nj}-s_n}\geq e^{\frac{1}{\varepsilon}},
    \end{equation*}
    because of the fact that $v_{n}\leq 1$, we can obtain:
    \begin{equation*}
        \|\frac{d\theta_t}{dt}-\frac{1}{N}\sum_{n} v_n\partial_\theta s_n\|_2\leq M(k-2)e^{-\frac{1}{\varepsilon}}.
    \end{equation*}
\end{proof}

\begin{theorem}
  Training Low-Rank DIQNN under cross-entropy loss and gradient flow algorithm, if the following three assumptions are satisfied in a small time interval $[t-\Delta t,t+\Delta t]$ for some $\Delta t>0$:
   \begin{itemize}
       \item $\|\Phi(x_n,\theta_t)\|_2=a_n\|\theta_t\|_2^L$
       \item The set $\{s_{nj}\}_{j=1,j\neq y_n}^{k}$ satisfy the $\varepsilon-$separated condition for some $\varepsilon>0$ ($\forall n\in [N]$), and $\{\|\partial_\theta s_{nj}\|_2\}_{n\in [N], j\in [k]}$ is uniformly bounded with some constant $M$.
       \item $c-1 \leq \frac{2m}{\sqrt{1-m^2}}$, where $m=\cos(\mathbf{a},\mathbf{v})>0$, $\mathbf{a}=(1/a_1,\dots,1/a_N)^T$
   \end{itemize}
   Then at time $t$ we have:
   \begin{equation*}
    \frac{d\Delta\mu}{dt}\geq -\frac{M^2(k-2)\|\mathbf{a}\|_1}{N\|\theta_t\|_2^L}e^{-\frac{1}{\varepsilon}}.
\end{equation*}
\end{theorem}

\begin{proof}
    It is obvious that $s_{n}=(\Phi(x_n,\theta))_{y_n}-(\Phi(x_n,\theta))_{j_n}$ which satisfy $s_n(a\theta)=a^Ls_n(\theta)$, then by Lemma 1 we know $\langle \partial_\theta s_{n}, \theta_t\rangle= Ls_n$, then we can decompose $\partial_\theta s_n$ into two orthogonal directions:
    \begin{equation*}
        \partial_\theta s_n=\frac{Ls_n}{\|\theta_t\|_2^2}\theta_t+\mathbf{\alpha_n}.
    \end{equation*}
    And using the first assumption, we can take the derivative w.r.t $t$ on both sides and then have: 
    \begin{equation*}
\langle \partial_{\theta}\|\Phi(x_n,\theta_t)\|_2,\frac{d\theta_t}{dt}\rangle=\langle\frac{L\|\Phi(x_n,\theta_t)\|_2}{\|\theta_t\|_2^2}\theta_t,\frac{d\theta_t}{dt}\rangle.
    \end{equation*}
    Then we can explicitly compute $\frac{d\Delta\mu}{dt}$:
    \begin{equation}\label{1}
        \begin{aligned}
        \frac{d}{dt}(\Delta\mu)=&\langle\frac{1}{N}\sum_{n}\frac{\|\Phi(x_n,\theta_t)\|_2\partial_\theta s_{n}-s_{n}\partial_\theta(\|\Phi(x_n,\theta_t)\|_2)}{\|\Phi(x_n,\theta_t)\|_2^2},\frac{d\theta_t}{dt}\rangle\\
        =&\langle\frac{1}{N}\sum_{n}\frac{\mathbf{\alpha_{n}}}{\|\Phi(x_n,\theta_t)\|_2},\frac{d\theta_t}{dt}\rangle\\
        =&\langle\frac{1}{N\|\theta_t\|_2^L}\sum_{n}\frac{\mathbf{\alpha_{n}}}{a_n},\frac{d\theta_t}{dt}\rangle.
        \end{aligned}
    \end{equation}
    And by the decomposition of $\partial_\theta s_n$, we have:
    \begin{equation*}
        \begin{aligned}
        \langle\frac{1}{N\|\theta_t\|_2^L}\sum_{n}\frac{\mathbf{\alpha_{n}}}{a_n}, \frac{1}{N}\sum_{n} v_n\partial_\theta s_n\rangle =& \frac{1}{N^2\|\theta_t\|_2^L}\langle\sum_n\frac{\mathbf{\alpha_n}}{a_n},\sum_n v_n\mathbf{\alpha_n}\rangle\\
        =& \frac{\mathbf{a}^TA\mathbf{v}}{N^2\|\theta_t\|_2^L}
        \end{aligned}.
    \end{equation*}
    Since $\cos(\mathbf{a},\mathbf{v})=m$, we can decompose $\mathbf{a}$: $\mathbf{a}=\frac{m\|a\|_2}{\|v\|_2}\mathbf{v}+\mathbf{b}$. Then we have: 
    \begin{equation*}
        \mathbf{a}^TA\mathbf{v}=\frac{\mathbf{v}^TA\mathbf{v}\|\mathbf{b}\|_2}{\|\mathbf{v}\|_2}(\frac{m}{\sqrt{1-m^2}}+\frac{\mathbf{\hat{b}}^TA\mathbf{\hat{v}}}{\mathbf{\hat{v}}^TA\mathbf{\hat{v}}}),
    \end{equation*}
    where $\mathbf{\hat{b}},\mathbf{\hat{v}}$ are unit vectors. Then we can do spectrum decomposition to the Gram matrix $A=\partial_\theta S\partial_\theta S^T-\frac{L^2}{\|\theta_t\|_2^2}SS^T=(\mathbf{\alpha_1},\mathbf{\alpha_2,\dots,\mathbf{\alpha_N}})^T(\mathbf{\alpha_1},\mathbf{\alpha_2,\dots,\mathbf{\alpha_N}})$: $A=Q^TDQ$, where $D=diag(\lambda_1,\lambda_2,\dots,\lambda_N)$ consist of eigenvalues of $A$, $Q$ is a orthogonal matrix, and we can assume $\lambda_1\geq \lambda_2\geq\cdots\geq \lambda_N\geq 0$, then we know $\mathbf{\hat{v}}^TA\mathbf{\hat{v}}\geq \lambda_N$. And since $\langle \mathbf{\hat{b}},\mathbf{\hat{v}}\rangle=0\Rightarrow\langle Q\mathbf{\hat{b}},Q\mathbf{\hat{v}}\rangle=0$, we can have:
    \begin{equation*}
        \mathbf{\hat{b}}^TA\mathbf{\hat{v}}=(Q\mathbf{\hat{b}})^T(D-\frac{\lambda_N+\lambda_1}{2} I)(Q\mathbf{\hat{v}})\geq \frac{\lambda_N-\lambda_1}{2}.
    \end{equation*}
    Then we get:  
    $\frac{\mathbf{\hat{b}}^TA\mathbf{\hat{v}}}{\mathbf{\hat{v}}^TA\mathbf{\hat{v}}}\geq \frac{\lambda_N-\lambda_1}{2\lambda_N}=\frac{1-c}{2}$. And by the third assumption, we have:
    \begin{equation*}
        \mathbf{a}^TA\mathbf{v}=\frac{\mathbf{v}^TA\mathbf{v}\|\mathbf{b}\|_2}{\|\mathbf{v}\|_2}(\frac{m}{\sqrt{1-m^2}}+\frac{\mathbf{\hat{b}}^TA\mathbf{\hat{v}}}{\mathbf{\hat{v}}^TA\mathbf{\hat{v}}})\geq \frac{\mathbf{v}^TA\mathbf{v}\|\mathbf{b}\|_2}{\|\mathbf{v}\|_2}(\frac{m}{\sqrt{1-m^2}}+\frac{1-c}{2})\geq 0.
    \end{equation*}
    Then we can get:
    \begin{equation*}
        \frac{d}{dt}(\Delta\mu)\geq \langle\frac{1}{N\|\theta_t\|_2^L}\sum_{n}\frac{\mathbf{\alpha_{n}}}{a_n}, \frac{d\theta_t}{dt}-\frac{1}{N}\sum_{n=1}^N v_n\partial_\theta s_n\rangle,
    \end{equation*}
    where 
    \begin{equation*}
        |\langle\frac{1}{N\|\theta_t\|_2^L}\sum_{n}\frac{\mathbf{\alpha_{n}}}{a_n}, \frac{d\theta_t}{dt}-\frac{1}{N}\sum_{n=1}^N v_n\partial_\theta s_n\rangle|\leq \frac{1}{N\|\theta_t\|_2^L}\|\sum_{n}\frac{\mathbf{\alpha_{n}}}{a_n}\|_2\cdot\|\frac{d\theta_t}{dt}-\frac{1}{N}\sum_{n=1}^N v_n\partial_\theta s_n\|_2.
    \end{equation*}
    Then by lemma 2 we can obtain:
    \begin{equation*}
    \begin{aligned}
    |\langle\frac{1}{N\|\theta_t\|_2^L}\sum_{n}\frac{\mathbf{\alpha_{n}}}{a_n}, \frac{d\theta_t}{dt}-\frac{1}{N}\sum_{n=1}^N v_n\partial_\theta s_n\rangle|&\leq \frac{1}{N\|\theta_t\|_2^L}\sum_{n}\|\frac{\partial_\theta s_n}{a_n}\|_2M(k-2)e^{-\frac{1}{\varepsilon}}\\
    &\leq \frac{M^2(k-2)\|\mathbf{a}\|_1}{N\|\theta_t\|_2^L}e^{-\frac{1}{\varepsilon}}.
    \end{aligned}
    \end{equation*}
    Finally, we obtain the result in the theorem:
    \begin{equation*}
    \frac{d\Delta\mu}{dt}\geq -\frac{M^2(k-2)\|\mathbf{a}\|_1}{N\|\theta_t\|_2^L}e^{-\frac{1}{\varepsilon}}.
\end{equation*}
\end{proof}

We observe that $k=2$ for binary classification task, therefore, the above inequality becomes: $\frac{d}{dt}(\Delta\mu)\geq 0$, which is consistent with the theorem in solving XOR problem.

\begin{figure}[htbp!]
    \centering
    \subfigure{\includegraphics[width=0.23\textwidth]{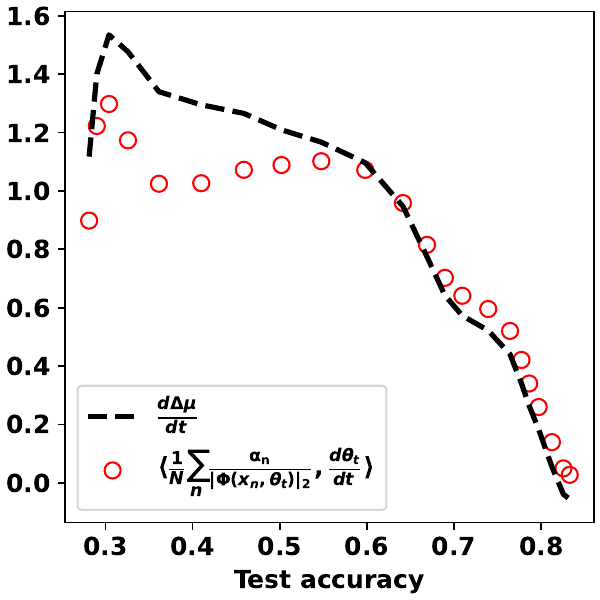}} 
    \subfigure{\includegraphics[width=0.23\textwidth]{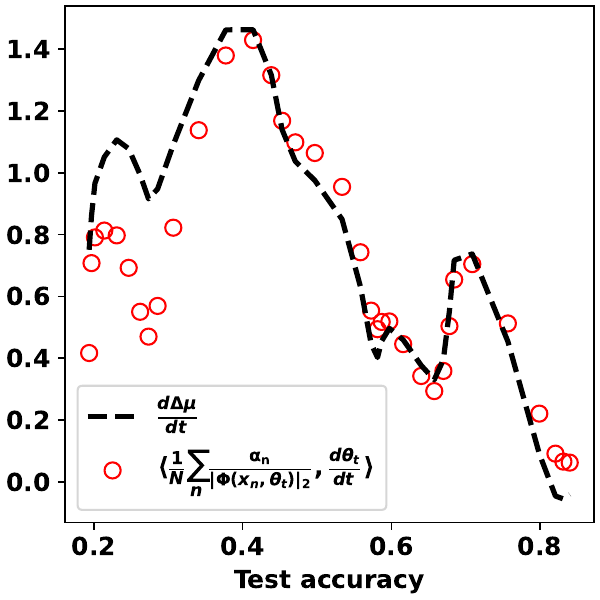}} 
    \subfigure{\includegraphics[width=0.23\textwidth]{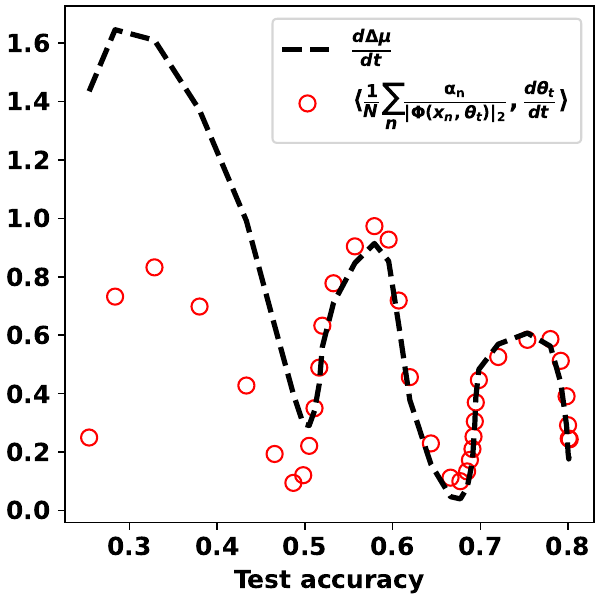}} 
    \subfigure{\includegraphics[width=0.23\textwidth]{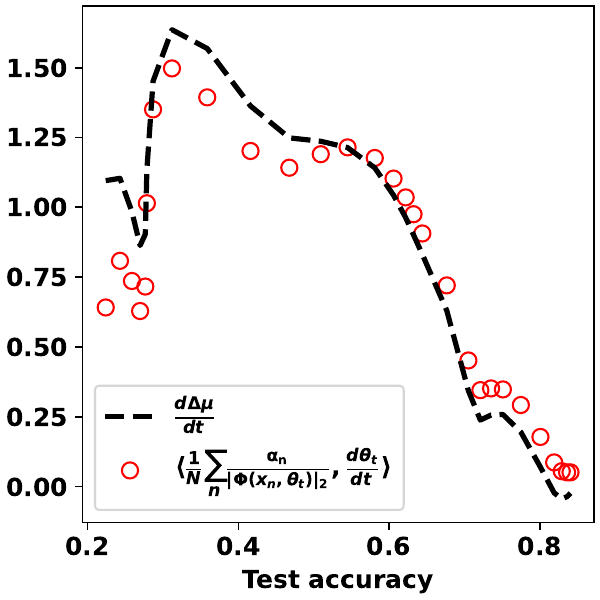}} 
    \subfigure{\includegraphics[width=0.23\textwidth]{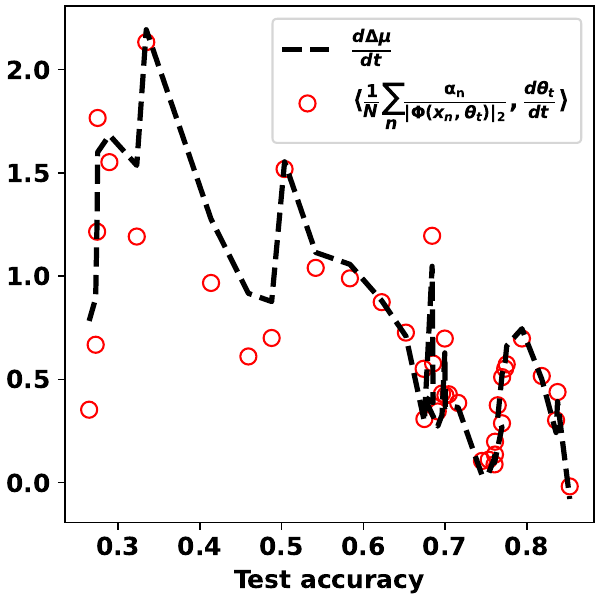}} 
    \subfigure{\includegraphics[width=0.23\textwidth]{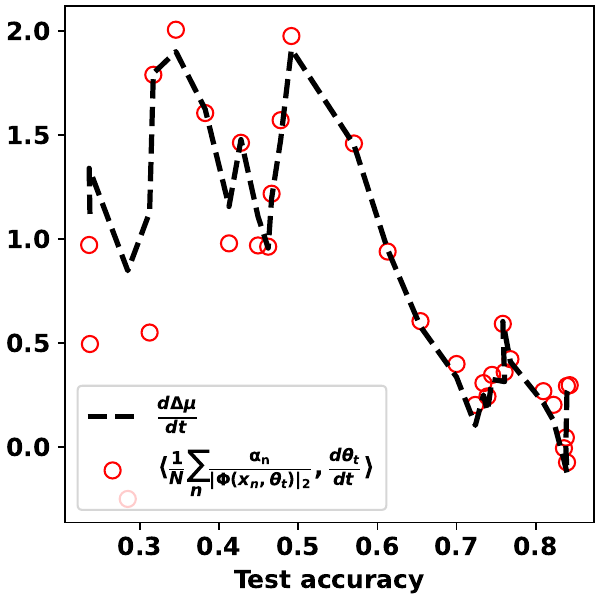}} 
    \subfigure{\includegraphics[width=0.23\textwidth]{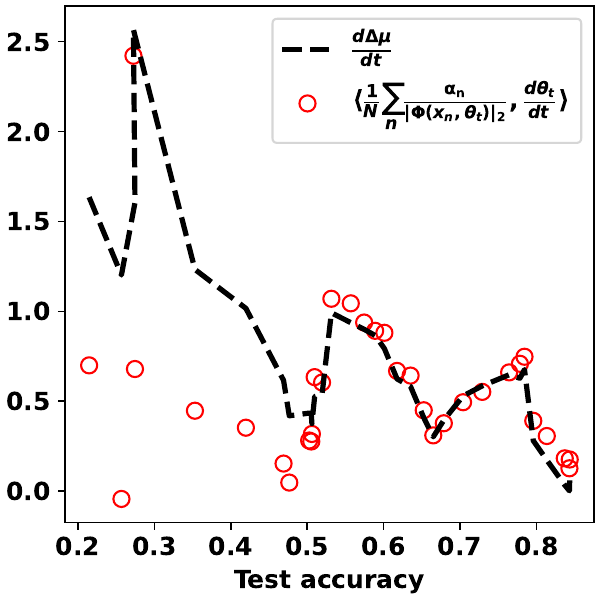}} 
    \subfigure{\includegraphics[width=0.23\textwidth]{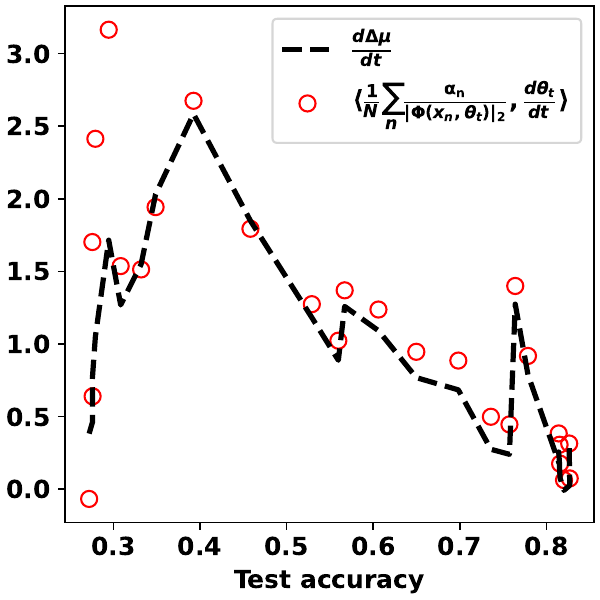}} 
    \caption{Verification of equality:  $\frac{d}{dt}(\Delta\mu)=\langle\frac{1}{N}\sum_{n}\frac{\mathbf{\alpha_{n}}}{\|\Phi(x_n,\theta_t)\|_2},\frac{d\theta_t}{dt}\rangle$. Each column represents the experiment under a certain random seed, and the first row refers to training with gradient descent (GD) algorithm, while the second row refers to training with stochastic gradient descent (SGD) algorithm.}
    \label{assumption_verification}
\end{figure}

Note that under the first assumption in our theorem, we obtain the following equality in our proof:
\begin{equation*}
\frac{d}{dt}(\Delta\mu)=\langle\frac{1}{N}\sum_{n}\frac{\mathbf{\alpha_{n}}}{\|\Phi(x_n,\theta_t)\|_2},\frac{d\theta_t}{dt}\rangle. 
\end{equation*} 
This equality can be verified by numerical experiments on MNIST throughout the training process of Low-Rank DIQNN (with one rank) under both GD and SGD, as shown in Figure~\ref{assumption_verification}. 

Finally, we point out that our theorem can be extended to models that satisfy the homogeneous property: 
\begin{equation*}
\exists \ L\ \ s.t.\ \ \Phi(x_n,\theta_t)=\|\theta_t\|_2^L\Phi(x_n,\frac{\theta_t}{\|\theta_t\|_2}),\ \forall a\in\mathbb{R},
\end{equation*}
e.g. DIQNN.

\medskip

